\documentclass[preprint,12pt]{elsarticle}

\usepackage{amssymb}
\usepackage{amsmath}

\usepackage{graphicx}
\usepackage{subcaption}
\usepackage{xcolor}
\definecolor{verde}{rgb}{0.0, 0.6, 0.0}
\definecolor{rojo}{rgb}{0.96, 0.76, 0.76}
\usepackage{adjustbox}
\usepackage{makecell}
\usepackage{multirow}
\usepackage{array}
\usepackage{xcolor}
\usepackage{comment}
\usepackage{algorithmic}
\usepackage[acronym]{glossaries} 
\usepackage{algorithm}
\usepackage{orcidlink}

\newcolumntype{C}[1]{>{\centering\arraybackslash}p{#1}}

\journal{Engineering Applications of Artificial Intelligence}

\newacronym{SERA}{SERA}{Single European Railway Area}
\newacronym{EU}{EU}{European Union}
\newacronym{IM}{IM}{Infrastructure Manager}
\newacronym[firstplural = Railway Undertakings (RUs)]{RU}{RU}{Railway Undertaking}

\begin{document}

\begin{frontmatter}



\title{Towards Equitable Rail Service Allocation Through Fairness-Oriented Timetabling in Liberalized Markets} 


\cortext[cor1]{Corresponding author}

\author[A]{David Muñoz-Valero \orcidlink{0000-0002-2509-9911}}
\ead{david.munoz@uclm.es}
\author[A]{Juan Moreno-Garcia \orcidlink{0000-0003-2430-145X}\corref{cor1}}
\ead{juan.moreno@uclm.es}
\author[B]{Julio Alberto López-Gómez \orcidlink{0000-0001-6291-6637}}
\ead{julioalberto.lopez@uclm.es}
\author[B]{Enrique Adrian Villarrubia-Martin \orcidlink{0009-0002-8006-5711}} \ead{enrique.villarrubia@uclm.es}

\address[A]{Escuela de Ingeniería Industrial y Aeroespacial de Toledo, Department of Technologies and Information Systems, Universidad de Castilla-La Mancha, Avenida Carlos III, s/n, Toledo, 45071, Spain}
\address[B]{Escuela Superior de Informática, Department of Technologies and Information Systems, Universidad de Castilla-La Mancha, Paseo de la Universidad 4, Ciudad Real, 13071, Spain}

\begin{abstract}
Over the last few decades, European rail transport has undergone major changes as part of the process of liberalization set out in European regulations. In this context of liberalization, railway undertakings compete with each other for the limited infrastructure capacity available to offer their rail services. The infrastructure manager is responsible for the equitable allocation of infrastructure between all companies in the market, which is essential to ensure the efficiency and sustainability of this competitive ecosystem. In this paper, a methodology based on Jain, Gini and Atkinson equity metrics is used to solve the rail service allocation problem in a liberalized railway market, analyzing the solutions obtained. The results show that the proposed methodology and the equity metrics used allow for equitable planning in different competitiveness scenarios. These results contrast with solutions where the objective of the infrastructure manager is to maximize its own profit, without regard for the equitable allocation of infrastructure. Therefore, the computational tests support the methodology and metrics used as a planning and decision support tool in a liberalized railway market.
\end{abstract}









\begin{keyword}
Liberalized railway market \sep Timetabling \sep Jain index \sep Gini coefficient \sep Atkinson equity index \sep Genetic algorithm


\end{keyword}


\end{frontmatter}

\section{Introduction}
Rail transport is one of the main drivers of a country's development and growth, due to its impact on communications, industry, trade, and tourism, among many other areas. Since 2001, the \gls{EU} has committed to a transformation of the railway market in its member countries by promoting a series of regulations (called railway packages) with the aim of creating a \gls{SERA}\cite{COM2011}.

Carrying out this objective is not a trivial task \cite{Ait2021, Yao2024Bi-objectiveLine}. However, there are many potential benefits to be derived from the creation of \gls{SERA}, which will improve rail supply in \gls{EU}, offer more competitive prices, and eliminate borders between countries by improving passenger and freight communication.

The first step to building \gls{SERA} is the liberalization of national railway markets. Traditionally, European railway markets presented a vertical structure comprising the \gls{IM} and a single \gls{RU}. On the one hand, \gls{IM} is in charge of managing the railway infrastructure, carrying out tasks such as assigning capacity to \gls{RU}, charging for track access and infrastructure maintenance, among others \cite{Pena2015}. On the other hand, \gls{RU} is the company that offers rail services to passengers using the infrastructure managed by \gls{IM}. In this traditional structure, \gls{RU} assumed a captive demand and there were no problems in managing infrastructure capacity and service schedules \cite{Besinovic2024}. When the market has a single actor with a monopoly of the system and plays the roles of \gls{IM} and \glspl{RU}, which was customary in many \gls{EU} countries, it is said to be a vertically and horizontally integrated market structure.

The liberalization of the national railway markets is a key step in the creation of \gls{SERA}. The new market structure will be composed of two different and separated entities: the \gls{IM} and multiple \glspl{RU}. The main feature of a liberalized railway market is competition. In this kind of market, the \glspl{RU} compete among themselves, both for the exploitation of the infrastructure offered by the \gls{IM}, and for the attraction of passengers to the services offered. This new context of competitive markets will allow the \glspl{RU} to provide their services throughout the \gls{EU} \cite{Caramello2017}, which will open up the passenger transport market to new \glspl{RU}, offering their services over the infrastructures managed by the \gls{IM}, and will also provide a greater and more varied range of services to passengers, together with more competitive prices \cite{Caramello2017}.

In a liberalized railway market, it is essential to optimize the capacity and use of the infrastructure offered by the \gls{IM} \cite{Smoliner2018}. The dynamics of a liberalized railway market are as follow: each \gls{RU} sends a set of requests to the \gls{IM}. A request consists of a time slot and a set of tracks on which the \gls{RU} is interested in offering a rail service. Once the requests have been received by the \gls{IM}, it is responsible for obtaining a schedule of rail services. It is not a trivial problem, since the \glspl{RU} compete with each other for infrastructure, and conflicts can appear. Formally, a conflict happens when two or more \glspl{RU} apply for access to the same track at the same time \cite{Schlechte2012}. In this way, the aim of \gls{IM} is, given the requests of the different \glspl{RU} in the market, to obtain a conflict-free rail service timetable. 

Two main approaches can be considered by the \gls{IM} when addressing the rail service allocation problem in liberalized markets: a profit-based approach and an equity-based approach. When a profit-based strategy is adopted, the aim of the \gls{IM} is to maximize its own profit from the \glspl{RU}. In this scenario, although it is true that the benefit to the \gls{IM} prevails, this strategy could mean prioritizing those willing to pay more for rail access or those that can generate more revenue through their services. Ultimately, this approach may lead to a concentration of resources in a few large and consolidated \glspl{RU}, limiting the entry of new competitors and ultimately leading the market to a monopoly situation. Moreover, when an equity-based approach \cite{Shao2022Equity-orientedSystem} is adopted, the aim is to distribute infrastructure capacity fairly among all the \glspl{RU}, regardless of their size or economic power. This approach has a number of advantages: by treating all the \glspl{RU} equally, it encourages new entrants to enter the market, which increases competition. Furthermore, greater competition can lead to more efficient use of infrastructure and improved quality of service. In addition, with more \glspl{RU} operating, prices for passengers may become more competitive, benefiting end users.

This paper proposes a framework to deal with the railway timetabling problem using equity measures to achieve a fair distribution among the \glspl{RU} competing for the infrastructure managed by the \gls{IM}. This study focuses, firstly, on identifying and applying equity criteria that will ensure a balanced distribution of rail capacity, allowing all the \glspl{RU} to have fair access to the infrastructure. Secondly, this paper develops a heuristic that takes fairness into account when making the allocation. The methods used in this article and the solutions obtained will allow a sustainable competitive ecosystem to be fostered, where multiple \glspl{RU} can operate efficiently, thus benefiting both operators and passengers with better prices and services.

The rest of the paper is structured as follows: Section \ref{sec:background} reviews a wide set of recent works showing the progress made in the European Union with regard to railway liberalization and the establishment of \gls{SERA}. Then, Section \ref{sec:Methodology} describes the methodology employed in this paper, formulating the elements of a liberalized market, defining the equity measures used and developing a heuristic designed to allocate rail services. Later, Section \ref{sec:Tests} describes the experiments and results obtained, showing the performance of the selected equity metrics in different scenarios, and analyzing the results of the allocation. Finally, Section \ref{sec:Conclusions} sets out some conclusions and further work derived from this paper.

\section{Background}\label{sec:background}

This section contains a literature review on the articles related to the objective presented in this work.

Gestrelius et al. \cite{Gestrelius2020} present a study focused on the perspective of practitioners working within Trafikverket, the Swedish railway infrastructure manager.  The study is based on seven key aspects of timetable quality, with particular relevance to competition management and the fulfillment of capacity requests. To carry out the study, a series of interviews were conducted with timetable designers at Trafikverket’s \gls{IM}, revealing several challenges and important conclusions regarding these aspects. A lack of clear guidelines and support for timetable constructors in handling competition was identified. For instance, one interviewee stated that ``competition management is a little bit hard because we have not been given any guidelines, and that is what is missing, so it’s up to each and every one to decide how to handle it''. The study concludes that the absence of guidelines leads to subjective decisions and inconsistent competition management. 

Regarding the fulfillment of capacity requests-often backed by legal documents and defended by the applicants-this should be a primary goal of the capacity allocation process. However, the previous study also highlights that a unilateral focus on satisfying requests may lead to network congestion and a decline in other aspects of timetable quality. As one respondent pointed out: ``it’s hard, because often... even though you follow all rules... you maybe feel that this is not going to be robust, this is not good, and the only thing you can do is to say it, but you have no means at all to do something about it''. These findings emphasize the need to develop clear guidelines and support tools to improve both competition management and request fulfillment in railway timetable planning. Furthermore, decision-support tools that incorporate such guidelines are necessary to assist planners in their tasks.

Additionally, recent research has addressed the impact of demand uncertainty on timetable quality. Cao and Feng \cite{Cao2020} propose a two-stage stochastic integer programming model for optimal capacity allocation in high-speed rail networks under random passenger demands. Their work emphasizes the need for heuristic and metaheuristic approaches to generate efficient and practical timetable solutions in complex operational environments.

As highlighted by Bruzzone et al. \cite{Bruzzone2023}, equity in transport can be understood from both distributive and procedural perspectives. Distributive equity concerns how transport-related benefits are shared among individuals, while procedural equity emphasizes the fairness of the decision-making processes that lead to such distributions. The study also clarifies the conceptual distinction between ``equity'' and ``equality'', which are frequently used interchangeably. While equality implies uniform treatment and access to the same resources or opportunities, equity acknowledges that individuals and groups may face different starting conditions, and thus requires differentiated measures to achieve comparable outcomes or levels of access. Thus, it is justified that equity in transportation is more related to the concepts of ``fairness'' and ``justice'' than to that of ``equality''.

Another important study \cite{Zhang2019} focuses on finding a balance between equity and efficiency in train timetabling. Although the study is set in the context of an urban transit line, the underlying problem is similar to that of capacity allocation in mainline railways, as both involve the fair and efficient distribution of limited resources among multiple users. In both settings, it is essential to balance operational efficiency with equity in the allocation of access to infrastructure. The study concludes that incorporating equity criteria helps ensure a fairer distribution of passenger waiting times. Moreover, it emphasizes the importance of implementing optimization tools that enable the analysis of different solutions that provide a trade-off between request fulfillment and other timetable quality criteria.

In terms of equity assessment, van Wee and Mouter \cite{Wee2021} provide a detailed overview of how transport equity can be evaluated in practice. They propose a framework that distinguishes between different equity types such as horizontal and vertical equity: while the former assumes equal treatment of individuals with similar needs, the latter advocates for differentiated treatment to account for social or economic disadvantage. The chapter further explores methodological approaches for measuring equity, including qualitative assessments and the use of quantitative indicators such as the Gini or Atkinson coefficients. While these indicators are valuable in assessing equity from a static perspective, optimization-based approaches are increasingly being adopted to handle the dynamic nature of transport systems. Zhang et al. \cite{Zhang2024} propose an approximate dynamic programming method for metro-train timetabling that incorporates passenger flow control and energy consumption, highlighting the role of data-driven and adaptive strategies in transport equity and efficiency.

The notion of fairness has gained attention in recent years, particularly in contexts where limited resources must be allocated among competing agents, such as in liberalized railway markets. Karsu and Morton \cite{Karsu2015} offer a comprehensive framework for incorporating inequity aversion into optimization models. The paper highlights various approaches to balance efficiency and fairness in decision-making processes. Li et al. \cite{Zhang2019} also investigate the trade-off between efficiency and fairness in the context of timetable design for a single urban rail transit line under time-dependent demand. Their study formulates a model that integrates both operator efficiency and user fairness, showing how various timetabling strategies can affect waiting times and train occupancy levels across passenger groups.

More recently, Luan et al. \cite{Luan2023} focus on traffic management optimization while explicitly modeling passenger route choice and using the Gini coefficient as a fairness indicator. Their approach introduces inequity-averse objectives to railway traffic management, aiming to reduce disparities in travel delays. Finally, Reynolds et al. \cite{Reynolds2023} examine fairness in timetable rescheduling scenarios where multiple train operating companies compete for infrastructure slots. Their work evaluates how different rescheduling strategies affect fairness between operators, using metrics derived from both allocation and performance outcomes. The findings underscore the importance of incorporating fairness explicitly into the timetable adjustment process, particularly in liberalized railway markets.

Although much of the literature has focused on railway systems, parallels can be drawn with research in other domains where resource allocation and scheduling are central challenges. In the context of cloud computing, Aron and Abraham \cite{Aron2022} offer a comprehensive survey of meta-heuristic and AI-based approaches for efficient resource scheduling under uncertainty and variability. These techniques, widely applied in computing, provide valuable insights for developing advanced railway scheduling tools in environments characterized by limited capacity and multiple competing agents. 

In summary, the aforementioned studies highlight the importance of balancing efficiency and equity in resource allocation and timetable planning, particularly in the context of competition management among operators. The implementation of equity criteria and optimization tools can contribute to the development of a fairer and more efficient system. In this way, this work is a contribution to filling the gap found in the literature review, allowing equity to be explicitly incorporated into the infrastructure allocation process and providing a decision support tool for infrastructure managers. 

\section{Methodology}\label{sec:Methodology}

As stated above, this work approaches the timetabling problem with a fairness-oriented perspective in the sense of equitable allocation between \glspl{RU} based on the agreed framework capacity of each RU. The formulation of the problem  and the different fairness indices used to pursue the fair capacity assignment are presented in Section \ref{sec:formulation}. Later, the proposal for generating a fairness-oriented timetable is introduced in Section \ref{sec:proposed_framework}.

\subsection{Formutation}\label{sec:formulation}

First, it will be shown how the \glspl{RU} have been mathematically formulated, their capabilities, and the way service requests are made. 

Equation \ref{ecu:RUs_definition} shows the representation used to model the \glspl{RU}.

\begin{equation}\label{ecu:RUs_definition}
    RU = (RU_1, \cdots, RU_n)
\end{equation}

Each RU has an assigned maximum capacity over the total infrastructure, as shown in Equation \ref{ecu:RUs_capacity}. 
    \begin{equation}\label{ecu:RUs_capacity}
        C = (c_1, \cdots, c_n )
    \end{equation}

Additionally, each \gls{RU} makes its requests to the \gls{IM} in the request process. To model this, the structure defined in Equation \ref{ecu:RUs_requests} is used.
\begin{equation}\label{ecu:RUs_requests}
    R = (R_1, \cdots, R_n)
\end{equation}

where each $R_i$ indicates the service request of each $RU_i$ and is formulated as shown in Equation \ref{ecu:RUs_service_i}.

\begin{equation}\label{ecu:RUs_service_i}
    R_i = ( (r_{i,|R_1|},w_{i,|R_1|}), \cdots, (r_{i,|R_i|},w_{i,|R_i|}) )
\end{equation}

where $r_{i,k}$ is the service $k$ requested by $RU_i$ and $w_{i,k}$ represents the importance that $RU_i$ assigns to the scheduling of service $r_{i,k}$, with the condition that $\sum^{|R_i|}_{k=1} w_{i,k} = 1$. The operator $|R_i|$ indicates the number of elements in $R_i$.

It can be seen that the proposal considers the importance of allowing the RU to indicate the services it is most interested in, assigned by using an importance percentage denoted as $w_{i,k}$. In this document, this percentage will be referred to as ``importance''. This percentage does not affect the general case where all services have the same importance for an RU, since in that case, each $w_{i,k}$ can be assigned to $\frac{1}{|R_i|}$.

The proposal for generating a fairness-oriented timetable involves scheduling considering fairness measures. To measure fairness, Jain's fairness index \cite{Jain1998}, the Gini coefficient \cite{Gini1912}, and the Atkinson index \cite{Atkinson1970} adapted to the problem have been used. Several recent studies have demonstrated the applicability of these indices in diverse transportation and communication contexts: Jain's index has been used to balance fairness and efficiency in wireless and vehicular networks \cite{Elnaby2021, Avcil2024}, the Gini coefficient has been used to assess spatial and temporal equity in hazardous material routing and public transport accessibility \cite{Romero2016, Raza2023}, while the Atkinson index has supported equity assessments in micromobility operations and infrastructure investment planning \cite{Lina2024, Chen2025}.The equations defining these measures will be detailed below.

Formula \ref{ecu:jain_equity_general} shows the general expression of Jain's fairness index. 
\begin{equation}\label{ecu:jain_equity_general}
    J = \frac{\left( \sum_{i=1}^{n} x_i \right)^2}{n \cdot \sum_{i=1}^{n} x_i^2}
\end{equation}

where $n$ is the total number of elements and $x_i$ the resources assigned to $i$.

It was proposed by Rajendra K. Jain in 1984 and is independent of population size, scale, and metric. The operation of Jain's index is designed to penalize unfair allocations by using the sum of squares in the denominator, which takes on a very high value when one user receives significantly more resources than others. Additionally, Jain's index obtains a normalized value, making it easier to interpret. This index has defined limits ranging from 0 to 1, where 1 indicates high fairness and 0 indicates unfairness.

An adaptation of this index has been made so that it can be used in this problem by changing the value of $x_i$. Specifically, two ideas have been introduced in the new value used. The first is that the original equation sums the resources granted; in this work, the importance of the services granted by summing these has been used. On the other hand, the sum of the obtained importances is raised to a value $\alpha$ to penalize sums with lower values (Equation \ref{eq:x_i_jain_alpha}). This exponent has been termed ``sensitivity''.

\begin{equation}
x_i = \left(\sum_{j=1}^{m} w_{ij}\right)^\alpha
\label{eq:x_i_jain_alpha}
\end{equation}

To explain the functioning of this formula, an example will be shown. Suppose that services between two RUs are to be planned, with the maximum number of services to be planned being $N_{serv} = 7$, and the capacities $C = (c_1, c_2) = (3,4)$. Let the following service request be $S = (S_1,S_2)$, where:
\begin{itemize}
    \item $R_1 = ( (r_{1,1},w_{1,1}),(r_{1,2},w_{1,2}),(r_{1,3},w_{1,3})) = ( (r_{1,1},0.2),(r_{1,2},0.3),(r_{1,3},0.5))$.
    \item $R_2 = ( (r_{2,1},w_{2,1}),(r_{2,2},w_{2,2}),(r_{2,3},w_{2,3}),(r_{2,4},w_{2,4})) = ( (r_{2,1},0.45),(r_{2,1},0.25),$ \ 
    $(r_{2,3},0.2),(r_{2,4},0.1))$.
\end{itemize}

Additionally, it is assumed that the services granted to $RU_1$ and $RU_2$ from the requested services are $P_1 = (1,0,1)$ and $P_2 = (1,1,1,0)$, that is, $RU_1$ will be able to plan its services 1 and 3, while $RU_2$ will be able to schedule its services from 1 to 3, inclusive. To calculate Jain's Equation, the values of $I_1$ and $I_2$ will first be calculated, using a value $\alpha=10$ (Equation \ref{eq:x_i_jain_alpha}):
\begin{itemize}
    \item $I_1 = (0.2+0.5)^{10} = 0.7^{10} = 0.028$
    \item $I_2 = (0.45+0.25+0.2)^{10} = 0.9^{10} = 0.349$
\end{itemize}

Then, applying Equation \ref{ecu:jain_equity_general}, the following result:

\begin{quote}
    $ J = \frac{(0.028+0.349)^2}{2 \cdot (0.028^2+0.349^2)}= \frac{0.377^2}{2 \cdot (0.000784+0.121801)} = \frac{0.142}{2\cdot0.123} = 0.557 $
\end{quote}

It can be seen that it is an equity index because $RU_1$ has only received $70\%$ of its requests, while $RU_2$ has received $90\%$ of its requests, that is, it is an unfair distribution.

To show the effect of $\alpha$ in the Equation, the result will be shown by applying $\alpha=1$. Equation \ref{ecu:jain_equity_general} will return the value:
\begin{itemize}
    \item $I_1 = (0.2+0.5)^{1} = 0.7$
    \item $I_2 = (0.45+0.25+0.2)^{1} = 0.9$
\end{itemize}

Then, applying Equation \ref{ecu:jain_equity_general}, the following result is obtained:

$$J = \frac{(0.7+0.9)^2}{n\cdot(0.7^2+0.9^2)} = \frac{1.3^2}{2\cdot(0.49+0.81)} = \frac{2.56}{2.6} = 0.985 $$

Recall that Jain's index penalizes inequitable allocations by using the sum of squares in the denominator. Thus, when a resource has been highly favored, its square becomes very high compared to the denominator. The use of low $\alpha$ values due to the normalization of importance causes this effect to cease to work. Using high $\alpha$ values solves this problem. It can be observed that with $\alpha=1$, a very high value, close to 1, the maximum, is obtained, despite it being an unfair distribution.

Equation \ref{ecu:gini_equity_general} shows the general expression of the Gini coefficient. 
\begin{equation}\label{ecu:gini_equity_general}
    G = \frac{\sum_{i=1}^{n} \sum_{j=1}^{n} |x_i - x_j|}{2 \cdot n^2 \cdot \bar{x}}
\end{equation}

where $n$ is the total number of elements, $x_i$ are the resources allocated to $i$ and $ \bar{x} $ is the arithmetic mean of the resources (i.e., the total resources divided by 
$n$).

\begin{figure}[!ht]
    \centering
    \includegraphics[width=0.5\textwidth]{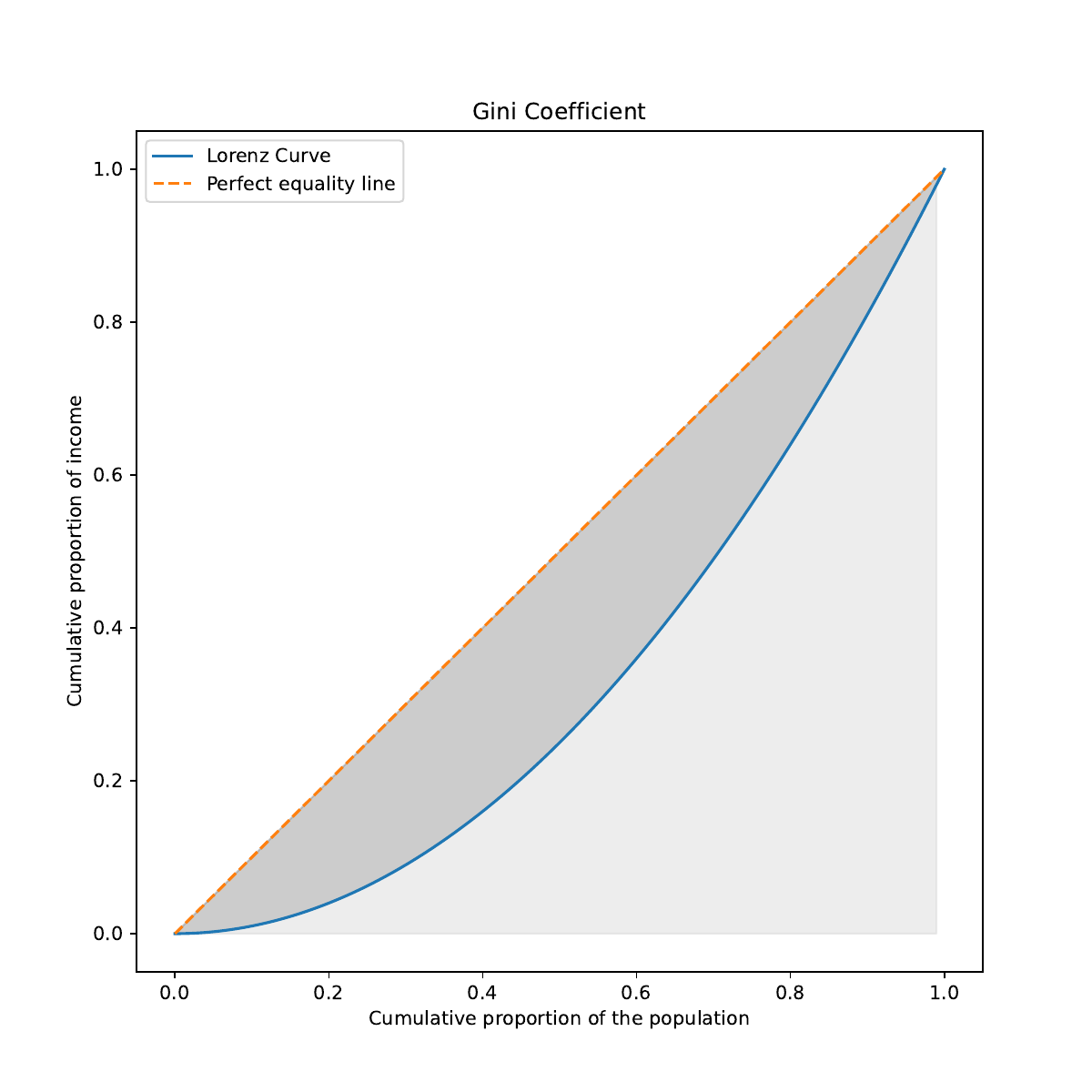}
    \caption{Calculation of the Gini coefficient.}
    \label{fig:area_gini_coefficient}
\end{figure}

Corrado Gini introduced the so-called ``Gini coefficient" in 1912, which is a method to show income inequality within a population. For this, this measure takes values between 0 and 1, with 0 being perfect equality (everyone has the same income) and 1 being perfect inequality (one individual concentrates all the income). Its operation is based on the Lorenz curve, which shows the cumulative distribution of income. In the case of perfect income equality, the Lorenz curve consists of a straight diagonal line, representing the line of perfect equality. Figure \ref{fig:area_gini_coefficient} shows the operation of the Gini coefficient, which calculates the percentage of the area of the curve (dark gray) with respect to the total area under the line of perfect equality (light gray).

In this case, the Gini Coefficient Equation had to be adapted to the problem. Specifically, the sum of the importances has been used in the form shown in Equation \ref{eq:x_i_jain_alpha} for the values of $x_i$ and $x_j$. The motivation is the same as in the case of Jain's Equation. Finally, Equation \ref{eq:fairness_gini} shows the equity measure that uses the Gini coefficient adapted to this problem. Since the best value of the Gini Coefficient is 0, the obtained value must be inverted so that the best equity measure is 1.

\begin{equation}
\text{Fairness} = 1 - G
\label{eq:fairness_gini}
\end{equation}

Next, the result obtained for the equity measure will be shown using the data from the previous Jain example. Recall that $I_1=0.028$ and $I_2=0.349$, applying Equation \ref{ecu:gini_equity_general}:
\begin{quote}
     $G = \frac{\sum_{i=1}^{n} \sum_{j=1}^{n} |x_i - x_j|}{2 \cdot n^2 \cdot \bar{x}} =  \frac{|0.028-0.028| + |0.028-0.349| + |0.349-0.349| + |0.349-0.028|}{2 \cdot 2^2 \cdot \frac{0.028+0.349}{2}} = \frac{0.642}{1.508} = 0.426$
\end{quote}

Then, the equity measure (Equation \ref{eq:fairness_gini}) returns $\text{Fairness} = 1 - G = 1 - 0.426 = 0.574$. An equity measure that indicates there is no equity in this case as it is far from 1.

The last of the equity measures tested is the Atkinson Equity Index. Anthony Barnes Atkinson worked on economic inequality, and his work has significantly contributed to economic and social policy. In 1970, he presented this index as a tool to measure income inequality in a more sensitive and adjustable manner. Equation \ref{ecu:atkinson_index} shows its expression. In this case, the sum of the importances for the value of $x_i$ is also used as in the previous cases (Equation \ref{eq:x_i_jain_alpha}).

\begin{equation}
A(\varepsilon) = 
\begin{cases} 
1 - \frac{1}{\mu} \left( \frac{1}{n} \sum_{i=1}^{n} x_i^{1-\varepsilon} \right)^{\frac{1}{1-\varepsilon}}  & \text{for } \varepsilon \neq 1, \infty \\
1 - \frac{1}{\mu} \left( \prod_{i=1}^{n} x_i \right)^{\frac{1}{n}} & \text{for } \varepsilon = 1 \\
1 - \frac{1}{\mu} \min(x_1, x_2, \ldots, x_n) & \text{for } \varepsilon = \infty
\end{cases}
\label{ecu:atkinson_index}
\end{equation}

where $n$ is the total number of elements and $x_i$ are the resources allocated to $i$.

The operation of the Equation is based on the inequality aversion parameter ($\epsilon$). The value of $\epsilon$ has been empirically adjusted to be suitable for evaluating equity in income distribution, highlighting sensitivity to differences in lower incomes to promote a more equitable distribution. The selected value is $\epsilon=0.5$. As in the case of Jain, the value obtained must be inverted to measure equity with high values (Equation \ref{eq:fairness_atkinson}).

\begin{equation}
\text{Fairness} = 1 - A
\label{eq:fairness_atkinson}
\end{equation}

Finally, using the data from the previous Jain example, the result obtained for the equity measure will be detailed. Recall that $x_1=0.028$ and $x_2=0.349$, applying Equation \ref{ecu:atkinson_index}, and considering that $\varepsilon=0.5$, the following result has been obtained:
\begin{quote}
     $A = 1 - \frac{1}{\mu} \left( \frac{1}{n} \sum_{i=1}^{n} x_i^{1-\varepsilon} \right)^{\frac{1}{1-\varepsilon}} = 1 - \frac{1}{\frac{0.028+0.349}{2}} \left( \frac{1}{2} \sum_{i=1}^{2} x_i^{1-0.5} \right)^{\frac{1}{1-0.5}} = 1 - \frac{1}{0.189} \left( \frac{1}{2} (0.028^{0.5}+0.349^{0.5}) \right)^{2} = 0.889 $
\end{quote}

Then, the equity measure (Equation \ref{eq:fairness_atkinson}) returns $\text{Fairness} = 1 - G = 1 - 0.889 = 0.110$. A value that indicates there is no equity in this case as it is very far from 1.

\begin{figure}[!ht]
    \centering
    \includegraphics[width=0.5\textwidth]{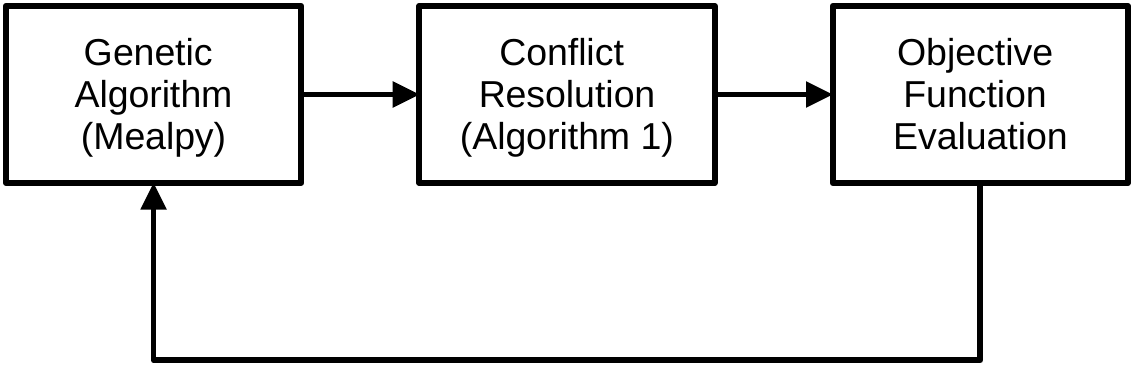}
    \caption{Generation of a Fairness-oriented Timetabling}
    \label{fig:algorithm_Fair-oriented_Timetabling}
\end{figure}

\subsection{Proposed framework}\label{sec:proposed_framework}

Once the measures to be used have been presented, the proposal for generating a fairness-oriented timetabling will be presented. The use of the equity measure will be introduced in two steps of the process: (1) in the objective function itself, and (2) in the process of eliminating conflicts between granted services, given that conflicts may arise in the planned schedule. The process for obtaining the final timetable consists of the following steps (Figure \ref{fig:algorithm_Fair-oriented_Timetabling}):
\begin{enumerate}

    \item Obtaining schedules. The use of genetic algorithms is proposed \cite{Holland1992} for schedule optimization. The Python library used is \textit{MEALPY} \cite{Vanthieu2023}, which offers a wide range of metaheuristic algorithms. This library includes the implementation of genetic algorithms, which will be used in this work.  
    \item Conflict resolution. The schedule generated in the previous step may contain conflicts that need to be resolved. Algorithm \ref{alg:conflicts_resolution} details the process followed to obtain a conflict-free schedule. Each of its actions is described below. First, all services that can be scheduled, i.e., those without conflicts, are planned (Line 1). Next, the services with conflicts are selected in $S_{with}$ (Line 2). Subsequently, a loop (Line 3) selects the most affected operator (Line 4), called $R_{affected}$, using the equity measure, planning the service that provides the highest equity (Line 5), and removing $S_{affected}$ and the services in conflict with it from $S_{with}$. The idea of the method is to perform actions that maintain the highest equity value. To achieve this, the most disadvantaged RU will be favored at each moment by granting it the service that best maintains the overall equity value.

\begin{algorithm}[!t]
\caption{Resolution of scheduling conflicts returned by the genetic algorithm.}
\label{alg:conflicts_resolution}
\begin{algorithmic}[1]
\REQUIRE Timetable generated by the genetic algorithm ($SC_{GA}$)
\ENSURE Conflict-free planned timetable ($SC_{without}$).
\STATE $Schedule(S_{without})$ \COMMENT{Schedule the services without conflicts}
\STATE $S_{with}$ $\leftarrow$ Obtain services with conflicts from $SC_{GA}$
\WHILE{$S_{with} \neq \varnothing$}
    \STATE $RU_{affected}$ $\leftarrow$ Find the most affected RU
    \STATE $Schedule(S_{affected})$ \COMMENT{Plan the service that provides the greatest improvement to equity ($S_{affected}$)}
    \STATE $S_{with} = S_{with} - S_{affected}$ \COMMENT{Remove $S_{affected}$ from $S_{with}$}
    \STATE $s_{with} = S_{with} - ServicesInConflict(S_{affected})$ \COMMENT{Remove the services that conflict with $S_{affected}$ from $S_{with}$}
\ENDWHILE
\end{algorithmic}
\end{algorithm}

    \item Evaluation of the objective function with the feasible schedule of planned services after the previous step. The objective function used is aimed at achieving an equitable distribution, also considering the benefit of the IM (Equation \ref{eq:objective_function}).

    \begin{equation}
        fitness = revenue \cdot fairness
        \label{eq:objective_function}
    \end{equation}

    The $revenue$ value is obtained by the sum of profit obtained by the scheduled services from each \gls{RU}, considering that deviations in the operational times proposed by the \gls{IM} lead to penalties in this profit.
    
    \item Since a genetic algorithm has been used, the two previous steps will be repeated until the desired number of epochs is reached. 
\end{enumerate}

\section{Experimental settings and results}\label{sec:Tests}

In the computational experience developed in this paper, a liberalized railway market with five RUs has been considered. In this context, three different scenarios regarding the framework capacities will be discussed: 

\begin{itemize}
    \item Unbalanced framework capacities: there are significant differences between the framework capacities of the RUs. The following framework capacities were considered: 55\%, 25\%, 10\%, 5\% and 5\%.
    \item Semi-balanced framework capacities: there are differences between the framework capacities of the RUs, but not as pronounced as in the previous case. The capacities considered are 30\%, 25\%, 20\%, 15\%, 10\%.
    \item Balanced framework capacities: each RU has a framework capacity of 20\%, ensuring that every RU can request the same number of services. This balanced configuration represents an ideal baseline, where the fair allocation is expected to yield minimal inequity (i.e., values approaching 0\% inequity indicate a near-perfect distribution).
\end{itemize}

The Genetic Algorithm from MEALPY (version $ =3.0.1 $) has been used for the optimization process, with cross-over probability $ = 0.95 $ and mutation probability $ = 0.025 $. The experiments were run in an Intel i7-13700KF CPU. The same seed has been used between runs of the different fairness indices for reproducibility. A total of 100 optimization epochs are use in each run. The software is publicly available in the following repository: \url{https://github.com/DavidMunozValero/GSA_M}.

Based on the framework capacities mentioned above, Table \ref{tab:requests} shows the number of requests made by each RU in the different cases.

\begin{table*}[htbp]
\centering
\begin{adjustbox}{max width=\textwidth}
\begin{tabular}{|l|c|c|c|}
\cline{2-4}
\multicolumn{1}{c|}{} & \textbf{Balanced} & \textbf{Semi-balanced} & \textbf{Unbalanced} \\
\cline{2-4} \hline
\textbf{RU1} & 10\% & 15\% & 28\% \\ \hline
\textbf{RU2} & 10\% & 12\% & 12\% \\ \hline
\textbf{RU3} & 10\% & 10\% & 5\% \\ \hline
\textbf{RU4} & 10\% & 8\% & 2\% \\ \hline
\textbf{RU5} & 10\% & 5\% & 2\% \\ \hline
\end{tabular}
\end{adjustbox}
\caption{Services requested by each RU based on different framework capacity scenarios.}
\label{tab:requests}
\end{table*}

\subsection{Impact of $\alpha$ on Fairness Metrics}

The hyperparameter $\alpha$ was introduced in Equation \ref{eq:x_i_jain_alpha} to allow the adjustment of the sensitivity of the different fairness indices. The transformation applied to the service allocation importance by raising them to the power $\alpha$ has a marked effect on the computed inequity values over training epochs. In particular, a higher $\alpha$ amplifies the differences among the importances. Therefore, five different $\alpha$ values have been tested in order to select the value which achieves the best results for each fairness index in the unbalanced framework capacity scenario: 1 (i.e., importances not modified), 5, 10, 25, and 50. This study regarding the $\alpha$ effect has been carried out considering unbalanced framework capacities for five different RUs, requesting a total of 50 services between them all. This study is conducted for the unbalanced framework because it is the case where randomness has the least effect due to the greater difference in capacities. Five runs of the algorithm are made for each $\alpha$ value.

In order to measure the performance of the proposed method regarding the fair distribution of resources, a normalized measure of inequity is introduced in Equation \ref{eq:inequity}. This value of inequity is understood as a normalized measure (percentage) of allocation differences between RUs based on the maximum difference value (e.g: considering two RUs, the maximum inequity would be to assign all services requested by the first,  $ I_1 = 1 $ and no service to the second one $ I_2 = 0 $).

\begin{equation}
\text{Inequity (\%)} = \frac{PD_{sum}}{IV_{max}} \cdot 100
\label{eq:inequity}
\end{equation}

where Equation \ref{eq:pairwise_diff} gives the sum of absolute differences between the assigned importance value for each pair of RUs ($PD_{sum}$), and Equation \ref{eq:max_ineq} defines the maximum inequity value ($IV_{sum}$) depending on the number of RUs (\(n\)), considering two cases: \(n\) is even or odd. 

\begin{equation}
PD_{sum} = \sum_{i=1}^{n} \sum_{j=1}^{n} \left| I_i - I_j \right|
\label{eq:pairwise_diff}
\end{equation}

where $I_i$ is the sum of the importances of $RU_i$.

\begin{equation}
IV_{max} =
\begin{cases}
\frac{n^2}{4}, & \text{if } n \text{ is even}\\[8pt]
\frac{n^2-1}{4}, & \text{if } n \text{ is odd}
\end{cases}
\label{eq:max_ineq}
\end{equation}

For instance, consider three RUs ($n=3$) with assigned importance values $I_1=0.78$, $I_2=0.23$, and $I_3=0.15$. According to Equation~\ref{eq:pairwise_diff}, the sum of the absolute differences between each pair is given by:
\begin{quote}
    $PD_{sum} = |I_1 - I_2| + |I_1 - I_3| + |I_2 - I_3| = |0.78 - 0.23| + |0.78 - 0.15| + |0.23 - 0.15| = 0.55 + 0.63 + 0.08 = 1.26$
\end{quote}

Since there is an odd number of RUs, the maximum inequity value in this case would be: 
$$ IV_{max} = \frac{n^{2} - 1}{4} = \frac{3^2 - 1}{4} = 2 $$

The effect of $\alpha$ on the fairness indices is shown in Table \ref{tab:inequity_summary_alpha} for Jain, Gini, and Atkinson fairness indices respectively. In this Table $\mu$ is the mean inequity obtained in the tests, and $\sigma$ is the standard deviation between the five different runs. In these tables, inequity is expressed as a percentage, where 100\% represents the worst possible allocation and 0\% corresponds to the ideal, perfectly fair distribution, based on Equation \ref{eq:inequity}. The most significant values (minimum mean inequity and standard deviation in this case) are highlighted in bold.

\begin{table*}[htbp]
\centering
\begin{adjustbox}{max width=\textwidth}
\begin{tabular}{|l|C{30pt}|C{80pt}|C{120pt}|}
\hline
\makecell{\textbf{Index}} & $\boldsymbol{\alpha} $ & \textbf{Inequity} ($ \mu \cdot 100 $) & \textbf{Inequity standard deviation} ($ \sigma \cdot 100 $) \\
\hline \hline
Jain     & 1  & 25.38 & 5.48 \\ \hline
Jain     & 5  & 17.85 & 1.57 \\ \hline
Jain     & 10 & 16.96 & 0.62 \\ \hline
Jain     & 25 & \textbf{16.89} & \textbf{0.14} \\ \hline
Jain     & 50 & 18.56 & 2.12 \\ \hline \hline
Gini     & 1  & 24.06 & 8.34 \\ \hline
Gini     & 5  & 17.55 & 0.80 \\ \hline
Gini     & 10 & \textbf{17.13} & \textbf{0.54} \\ \hline
Gini     & 25 & 22.95 & 13.20 \\ \hline
Gini     & 50 & 27.97 & 22.66 \\ \hline \hline
Atkinson & 1  & 26.23 & 5.73 \\ \hline
Atkinson & 5  & 20.81 & 3.27 \\ \hline
Atkinson & 10 & 18.64 & 2.75 \\ \hline
Atkinson & 25 & \textbf{17.17} & \textbf{0.56} \\ \hline
Atkinson & 50 & 17.44 & 0.78 \\ \hline
\end{tabular}
\end{adjustbox}
\caption{Inequity results with unbalanced capacities based on different $\alpha$ values.}
\label{tab:inequity_summary_alpha}
\end{table*}

\begin{figure}[!ht]
    \centering
    \begin{subfigure}[b]{0.32\textwidth}
        \centering
        \includegraphics[width=\textwidth]{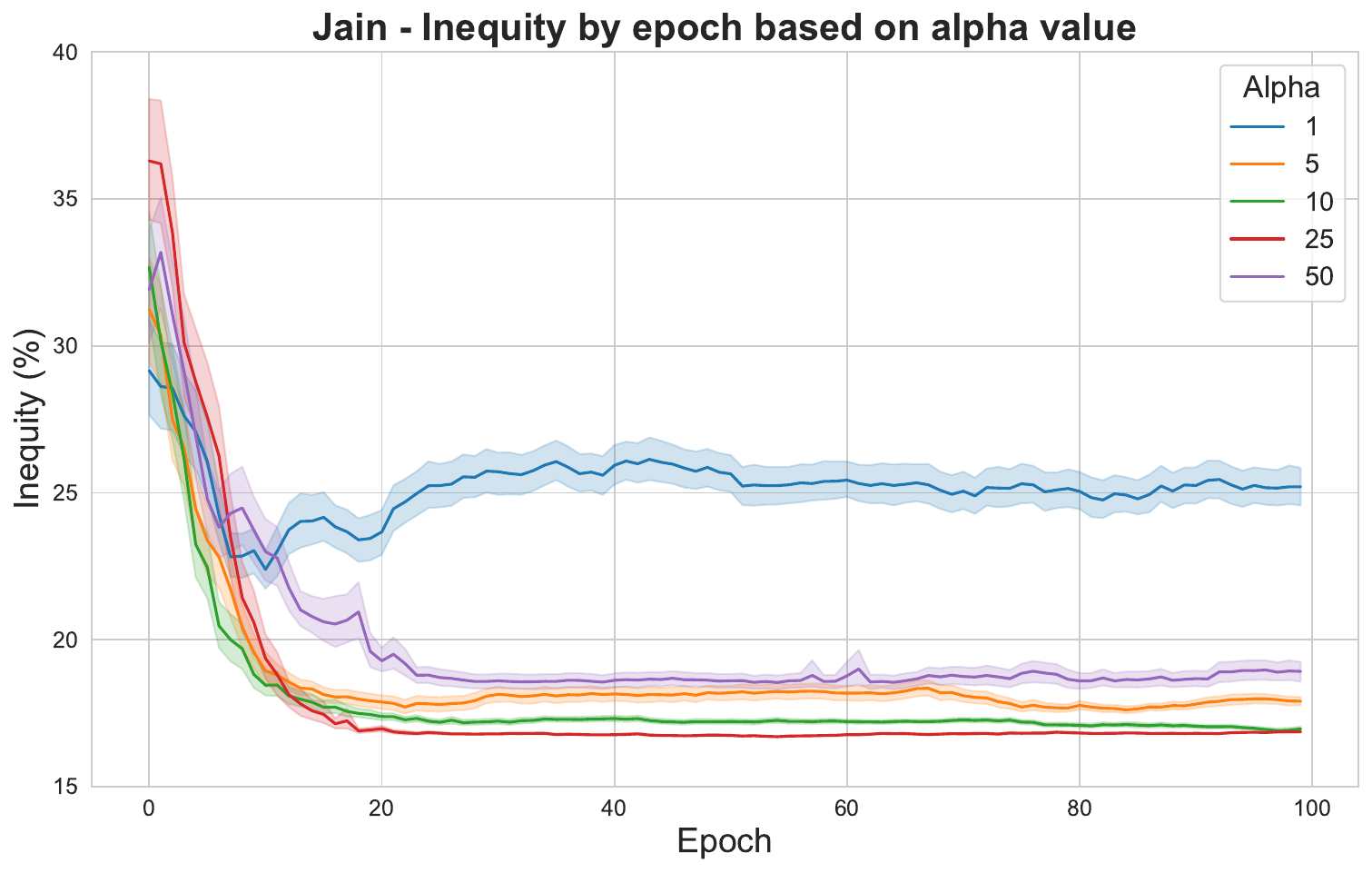}
        \caption{Jain}
        \label{fig:jain_inequity_alpha}
    \end{subfigure}
    \hfill
    \begin{subfigure}[b]{0.32\textwidth}
        \centering
        \includegraphics[width=\textwidth]{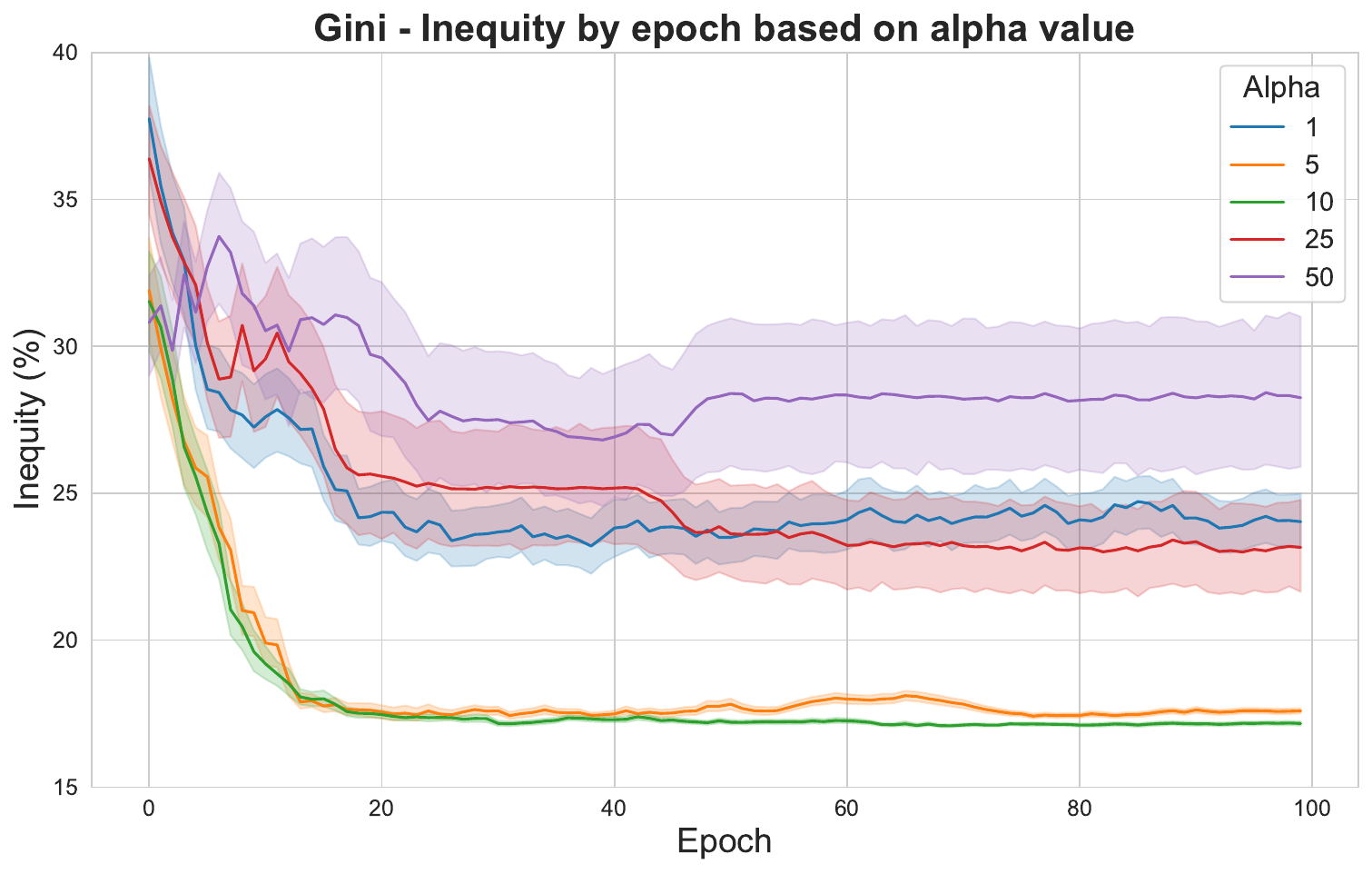}
        \caption{Gini}
        \label{fig:gini_inequity_alpha}
    \end{subfigure}
    \hfill
    \begin{subfigure}[b]{0.32\textwidth}
        \centering
        \includegraphics[width=\textwidth]{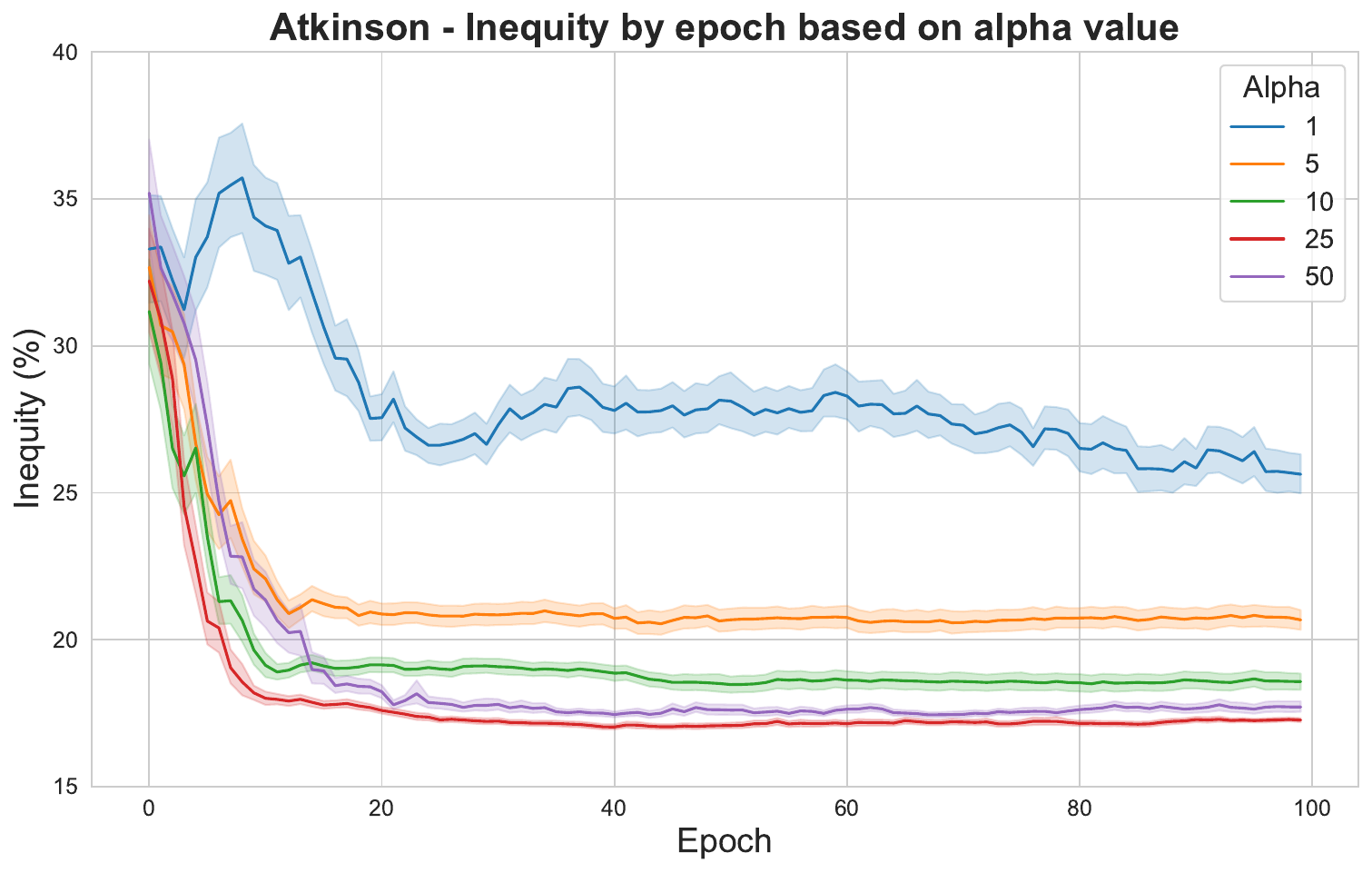}
        \caption{Atkinson}
        \label{fig:atkinson_inequity_alpha}
    \end{subfigure}
    
    \caption{Inequity (\%) evolution over training epochs.}
    \label{fig:inequity_alpha_comparison}
\end{figure}

Based on the observed results, for the Jain fairness index, an $\alpha$ value of 25 yields the lowest mean inequity and standard deviation. Similarly, the Atkinson fairness index also achieves optimal performance at $\alpha=25$, with both the minimum average inequity and the smallest dispersion observed. In contrast, the Gini fairness index attains the lowest overall inequity (and corresponding standard deviation) when $\alpha=10$, suggesting that the sensitivity of the Gini measure peaks at a lower exponent compared to the other indices. These trends are also evident in Figure \ref{fig:inequity_alpha_comparison}, where the mean inequity is plotted together with the standard deviation. 

For the remainder of this study, an $\alpha$ value of 25 will be used with the Jain and Atkinson indices, whereas an $\alpha$ value of 10 will be adopted for the Gini index. This choice is based on the observed performance in terms of both the average inequity and its variability.

\subsection{Testing different framework capacities scenarios}

Based on the previous values, a comparative study has been carried out in order to compare the effectiveness of the selected fairness indices in the three different framework capacity scenarios.

Starting with the unbalanced case, Table~\ref{tab:balanced_capacities} summarizes the performance metrics obtained under the balanced framework capacity scenario. The Table includes, for each run, the inequity, along with other key metrics such as: revenue, assigned importance for each RU, assigned importance (Equation \ref{eq:importance_percentage}) and assigned capacity (Equation \ref{eq:capacity_usage}). The difference between the two Equations mentioned before is that the assigned importance percentage (Equation \ref{eq:importance_percentage}) reflects solely the mean importance allocated to the RUs, whereas the assigned capacity (Equation \ref{eq:capacity_usage}) incorporates the specific capacity of each RU. If the framework capacity is identical across all RUs, both measures will be equivalent. The Table's results are grouped by fairness index and ranked by ascending inequity within each group.

\begin{equation}
P_I = \frac{I_{\text{total}}}{n} \cdot 100
\label{eq:importance_percentage}
\end{equation}

where: $ I_{\text{total}} = \sum_{i=1}^{n}{I_i}$.

\begin{equation}
C_{\text{usage}} = \sum_{i=1}^{n} I_i \cdot c_i
\label{eq:capacity_usage}
\end{equation}

\begin{table*}[htbp]
\centering
\begin{adjustbox}{max width=\textwidth}
\begin{tabular}{|c|c|c|c|c|c|c|c|c|c|c|c|}
\cline{2-12}
 \multicolumn{1}{c|}{} & \makecell{\textbf{Inequity} \\ \textbf{(\%)}} & \textbf{Run} & \textbf{Revenue} & \makecell{\textbf{Scheduled} \\ \textbf{Trains}} & \makecell{\textbf{$I_1$} \\ \textbf{(\%)}} & \makecell{\textbf{$I_2$}  \\ \textbf{(\%)}} & \makecell{\textbf{$I_3$}  \\ \textbf{(\%)}} & \makecell{\textbf{$I_4$} \\ \textbf{(\%)}} & \makecell{\textbf{$I_5$} \\ \textbf{(\%)}} & \makecell{\textbf{Assigned} \\ \textbf{Importance} \\ \textbf{(\%)}} & \makecell{\textbf{Assigned} \\ \textbf{Capacity} \\ \textbf{(\%)}} \\ \cline{2-12} \hline

\parbox[t]{2mm}{\multirow{5}{*}{\rotatebox[origin=c]{90}{\textbf{Jain}}}} & \textbf{0.89} & 2 & 3413.31 & 20 & 33.83 & 32.79 & 33.84 & 33.81 & 34.12 & 33.68 & 33.68 \\ 
& 0.96 & 1 & 2452.65 & 21 & 44.20 & 43.98 & 44.05 & 43.08 & 44.41 & 43.94 & 43.94 \\ 
& 1.23 & 5 & 3344.49 & 22 & 31.48 & 32.63 & 32.34 & 33.06 & 32.88 & 32.48 & 32.48 \\ 
& 1.29 & 4 & 2412.14 & 21 & 34.66 & 34.65 & 35.22 & 36.09 & 35.65 & 35.25 & 35.25 \\ 
& 4.38 & 3 & 2647.89 & 21 & 33.35 & 38.64 & 39.66 & 39.07 & 39.16 & 37.97 & 37.97 \\ \hline

\parbox[t]{2mm}{\multirow{5}{*}{\rotatebox[origin=c]{90}{\textbf{Gini}}}} & \textbf{1.23} & 5 & 2104.02 & 22 & 38.52 & 38.23 & 38.51 & 37.24 & 38.94 & 38.29 & 38.29 \\ 
& 1.91 & 3 & 2635.36 & 22 & 34.50 & 32.36 & 31.98 & 33.06 & 32.56 & 32.89 & 32.89 \\
& 2.01 & 2 & 2816.91 & 19 & 34.66 & 34.85 & 36.73 & 36.71 & 36.71 & 35.93 & 35.93 \\ 
& 2.33 & 1 & 2176.53 & 20 & 31.54 & 32.63 & 31.22 & 33.12 & 33.93 & 32.49 & 32.49 \\ 
& 2.70 & 4 & 2923.39 & 22 & 43.63 & 42.79 & 43.69 & 42.93 & 40.06 & 42.62 & 42.62 \\ \hline

\parbox[t]{2mm}{\multirow{5}{*}{\rotatebox[origin=c]{90}{\textbf{Atkinson}}}} & \textbf{1.39 } & 5 & 2487.46 & 21 & 37.73 & 38.23 & 38.51 & 37.24 & 38.94 & 38.13 & 38.13 \\
& 3.73 & 4 & 2613.38 & 22 & 41.42 & 40.71 & 38.01 & 42.52 & 39.25 & 40.38 & 40.38 \\
& 4.76 & 2 & 1233.40 & 21 & 29.04 & 35.43 & 33.90 & 34.66 & 33.18 & 33.24 & 33.24 \\ 
& 4.77 & 1 & 2889.62 & 20 & 40.81 & 41.79 & 40.56 & 35.97 & 38.15 & 39.46 & 39.46 \\ 
& 23.46 & 3 & 2721.45 & 22 & 46.53 & 50.56 & 51.58 & 49.43 & 18.41 & 43.30 & 43.30 \\ \hline

\parbox[t]{2mm}{\multirow{5}{*}{\rotatebox[origin=c]{90}{\textbf{Revenue}}}} & \textbf{21.25} & 1 & 6059.13 & 25 & 54.20 & 38.06 & 38.68 & 44.80 & 62.19 & 47.59 & 47.59 \\ 
& 28.34 & 5 & 6382.89 & 26 & 25.65 & 45.87 & 57.74 & 62.22 & 56.45 & 49.59 & 49.59 \\ 
& 29.28 & 3 & \textbf{6699.45} & 27 & 28.48 & 45.87 & 61.09 & 44.80 & 64.25 & 48.90 & 48.90 \\ 
& 34.61 & 2 & 6682.48 & \textbf{28} & 24.87 & 67.06 & 71.52 & 62.22 & 56.55 & \textbf{56.44} & \textbf{56.44} \\
& 35.57 & 4 & 5655.58 & 24 & 35.44 & 26.72 & 53.74 & 63.74 & 65.93 & 49.11 & 49.11 \\ \hline
\end{tabular}
\end{adjustbox}
\caption{Results by run for each fairness index. Balanced framework capacities.}
\label{tab:balanced_capacities}
\end{table*}

The columns $I_1$, $I_i$, ..., $I_5$ represent the assigned importance (expressed as a percentage) for each RU. Therefore, the smaller the differences between these values, the better the fairness. It can be observed that fairness-oriented approaches tend to produce solutions in which the assigned importance values for each RU are balanced. For example, in the second run of the Jain fairness index approach, the assigned importance values for all RUs ranged from 32.79\% to 34.12\%. Thus, the difference between the RU with the highest allocated resources and the one with the lowest is below 1.5\%.

In the semi-balanced capacity scenario (Table~\ref{tab:semi_balanced_capacities}), fairness-oriented approaches (Jain, Gini, and Atkinson) generally exhibit low inequity levels with nearly identical values for assigned importance and assigned capacity across most runs, indicating a homogeneous distribution of resources. The most significant values are highlighted in bold. However, one Atkinson run registers significantly high inequity (27.54\%), reflecting a higher dispersion. In contrast, revenue maximization yields the highest revenue values but at the expense of considerably higher inequity (ranging from 17.79\% to 43.15\%), resulting in a markedly unequal resource allocation.

Regarding the columns $I_1$, $I_i$, ..., $I_5$, a pattern similar to that observed earlier emerges: fairness-focused methods tend to yield outcomes in which the importance values assigned to each RU are evenly distributed. In this instance, the most equitable solution is achieved during the fifth run of the Gini coefficient method, where the importance values for all RUs ranged from 47.68\% to 48.85\%. Similarly, the difference between the RU with the highest allocated resources and the one with the lowest remains below 1.5\%.

\begin{table*}[htbp]
\centering
\begin{adjustbox}{max width=\textwidth}
\begin{tabular}{|c|c|c|c|c|c|c|c|c|c|c|c|}
\cline{2-12}
 \multicolumn{1}{c|}{} & \makecell{\textbf{Inequity} \\ \textbf{(\%)}} & \textbf{Run} & \textbf{Revenue} & \makecell{\textbf{Scheduled} \\ \textbf{Trains}} & \makecell{\textbf{$I_1$} \\ \textbf{(\%)}} & \makecell{\textbf{$I_2$}  \\ \textbf{(\%)}} & \makecell{\textbf{$I_3$}  \\ \textbf{(\%)}} & \makecell{\textbf{$I_4$} \\ \textbf{(\%)}} & \makecell{\textbf{$I_5$} \\ \textbf{(\%)}} & \makecell{\textbf{Assigned} \\ \textbf{Importance} \\ \textbf{(\%)}} & \makecell{\textbf{Assigned} \\ \textbf{Capacity} \\ \textbf{(\%)}} \\ \cline{2-12} \hline

\parbox[t]{2mm}{\multirow{5}{*}{\rotatebox[origin=c]{90}{\textbf{Jain}}}} &
\textbf{1.34} & 5 & 1288.70 & 23 & 47.06 & 47.68 & 46.74 & 46.13 & 47.67 & 47.05 & 47.06 \\
& 2.81 & 3 & 2414.82 & 26 & 49.76 & 45.83 & 49.29 & 48.75 & 49.33 & 48.59 & 48.52 \\
& 4.47 & 2 & 3483.19 & 27 & 52.65 & 52.98 & 52.07 & 53.93 & 47.67 & 51.86 & 52.32 \\
& 5.34 & 1 & 3855.54 & 25 & 56.24 & 62.41 & 63.05 & 62.95 & 63.93 & \textbf{61.71 }& \textbf{60.92} \\
& 12.09 & 4 & 1816.13 & 22 & 40.98 & 37.40 & 40.81 & 41.37 & 25.03 & 37.12 & 38.55 \\ \hline

\parbox[t]{2mm}{\multirow{5}{*}{\rotatebox[origin=c]{90}{\textbf{Gini}}}} & \textbf{1.02} & 5 & 3150.58 & 23 & 48.75 & 47.68 & 48.76 & 48.75 & 47.85 & 48.36 & 48.40 \\
& 1.32 & 3 & 3038.50 & 24 & 47.98 & 47.68 & 47.91 & 46.13 & 47.67 & 47.48 & 47.57 \\
& 1.84 & 1 & 3383.60 & 23 & 49.42 & 47.68 & 48.76 & 50.11 & 49.15 & 49.02 & 48.95 \\
& 2.39 & 4 & 2788.02 & 23 & 49.74 & 48.14 & 48.85 & 50.63 & 47.85 & 49.04 & 49.13 \\
& 3.07 & 2 & 3252.63 & 24 & 50.90 & 52.06 & 51.30 & 50.88 & 47.67 & 50.56 & 50.93 \\ \hline

\parbox[t]{2mm}{\multirow{5}{*}{\rotatebox[origin=c]{90}{\textbf{Atkinson}}}} & \textbf{1.81} & 1 & 3644.53 & 24 & 48.04 & 48.95 & 48.70 & 50.11 & 47.85 & 48.73 & 48.70 \\
& 2.85 & 3 & 3233.69 & 23 & 49.60 & 50.22 & 51.30 & 50.88 & 47.67 & 49.93 & 50.10 \\
& 3.82 & 4 & 3249.85 & 24 & 48.75 & 47.68 & 49.52 & 44.32 & 47.85 & 47.62 & 47.85 \\
& 4.40 & 5 & 3690.36 & 27 & 53.40 & 52.98 & 52.07 & 50.88 & 47.85 & 51.43 & 52.07 \\
& 27.54 & 2 & 2385.02 & 26 & 58.90 & 56.62 & 20.75 & 58.35 & 60.92 & 51.11 & 50.84 \\ \hline

\parbox[t]{2mm}{\multirow{5}{*}{\rotatebox[origin=c]{90}{\textbf{Revenue}}}} & \textbf{17.79} & 5 & 7210.23 & \textbf{29} & 56.63 & 50.70 & 47.30 & 71.02 & 50.85 & 55.30 & 55.07 \\
& 18.36 & 3 & 6356.56 & 27 & 49.62 & 62.54 & 63.11 & 71.02 & 50.85 & 59.43 & 58.96 \\
& 18.84 & 2 & \textbf{7258.89} & 27 & 56.11 & 59.78 & 47.24 & 71.02 & 50.85 & 57.00 & 57.08 \\
& 32.15 & 4 & 6625.12 & 27 & 46.79 & 55.24 & 49.88 & 66.43 & 86.75 & 61.02 & 56.58 \\
& 43.15 & 1 & 7244.70 & \textbf{29} & 73.54 & 59.78 & 20.69 & 74.08 & 50.85 & 55.79 & 57.48 \\ \hline
\end{tabular}
\end{adjustbox}
\caption{Results by run for each fairness index. Semi-balanced capacities.}
\label{tab:semi_balanced_capacities}
\end{table*}

In the unbalanced capacity scenario (Table~\ref{tab:unbalanced_capacities}), fairness-driven optimizations (Jain, Gini, and Atkinson) generally achieve consistent inequity levels around 16–17\% (with one Atkinson run at 19.52\%), and the assigned importance closely mirrors the assigned capacity, indicating a proportional allocation with respect to RU-specific capacities. In contrast, revenue maximization delivers the highest fitness and revenue values but incurs extremely high inequity (ranging from 75.86\% to 78.19\%), reflecting a markedly skewed distribution of resources between RUs. 

Regarding the columns $I_1$, $I_i$, ..., $I_5$, it is once again evident that fairness-centric methods yield outcomes where the importance values assigned to each RU are far more evenly distributed than those produced by the revenue maximization strategy. However, it is important to note that balancing this scenario of disparate framework capacities is considerably more challenging than in the previous two cases. In this instance, the most balanced solution was obtained during the fourth run of the Atkinson coefficient method, where the importance values for all RUs ranged from 25.67\% to 49.24\%. Conversely, the most balanced result using the revenue maximization approach reveals an inequity of 75.86\%, with values ranging from 11.78\% to 100\%. It is also important to note that even with fairness-oriented methods, some RUs may receive a significantly lower allocation if they have very few feasible requests; for instance, if one RU only requests two services and one turns out to be unfeasible to schedule, its final scheduled importance will be much lower than that of the other RUs. This is the case with $x_4$.

\begin{table*}[htbp]
\centering
\begin{adjustbox}{max width=\textwidth}
\begin{tabular}{|c|c|c|c|c|c|c|c|c|c|c|c|}
\cline{2-12}
 \multicolumn{1}{c|}{} & \makecell{\textbf{Inequity} \\ \textbf{(\%)}} & \textbf{Run} & \textbf{Revenue} & \makecell{\textbf{Scheduled} \\ \textbf{Trains}} & \makecell{\textbf{$I_1$} \\ \textbf{(\%)}} & \makecell{\textbf{$I_2$}  \\ \textbf{(\%)}} & \makecell{\textbf{$I_3$}  \\ \textbf{(\%)}} & \makecell{\textbf{$I_4$} \\ \textbf{(\%)}} & \makecell{\textbf{$I_5$} \\ \textbf{(\%)}} & \makecell{\textbf{Assigned} \\ \textbf{Importance} \\ \textbf{(\%)}} & \makecell{\textbf{Assigned} \\ \textbf{Capacity} \\ \textbf{(\%)}} \\ \cline{2-12} \hline

\parbox[t]{2mm}{\multirow{5}{*}{\rotatebox[origin=c]{90}{\textbf{Jain}}}} & \textbf{16.71}     & 5 & 3926.41 & 23 & 48.93 & 48.75 & 45.94 & 25.67 & 49.24 & 43.70 & 47.64 \\
& 16.78 & 2 & 3840.79 & 24 & 48.78 & 49.12 & 45.94 & 25.67 & 49.24 & 43.75 & 47.65 \\
& 16.90 & 3 & 3478.70 & 22 & 48.76 & 49.37 & 45.94 & 25.67 & 49.24 & 43.80 & 47.70 \\
& 17.03 & 1 & 3586.79 & 24 & 49.56 & 49.12 & 45.94 & 25.67 & 49.24 & 43.91 & 48.09 \\
& 17.23 & 4 & 3166.76 & 23 & 49.76 & 49.43 & 45.94 & 25.67 & 49.24 & 44.01 & 48.28 \\ \hline

\parbox[t]{2mm}{\multirow{5}{*}{\rotatebox[origin=c]{90}{\textbf{Gini}}}} & \textbf{16.58} & 5 & 4708.59 & 24 & 48.52 & 47.70 & 45.94 & 25.67 & 49.24 & 43.41 & 47.15 \\
& 16.69 & 4 & 3677.71 & 23 & 48.85 & 48.84 & 45.94 & 25.67 & 49.24 & 43.70 & 47.61 \\
& 17.16 & 3 & 5206.93 & 25 & 49.76 & 48.90 & 45.94 & 25.67 & 49.24 & 43.90 & 48.15 \\
& 17.30 & 1 & 3340.72 & 25 & 49.74 & 49.67 & 45.94 & 25.67 & 49.24 & 44.05 & 48.33 \\
& 17.94 & 2 & 4547.15 & 25 & 50.70 & 49.67 & 45.94 & 25.67 & 49.24 & 44.24 & 48.88 \\ \hline

\parbox[t]{2mm}{\multirow{5}{*}{\rotatebox[origin=c]{90}{\textbf{Atkinson}}}} & \textbf{16.49} & 4 & 4074.92 & 22 & 47.68 & 48.28 & 45.94 & 25.67 & 49.24 & 43.36 & 46.81 \\
& 16.63 & 5 & 4075.33 & 23 & 48.18 & 48.68 & 45.94 & 25.67 & 49.24 & 43.54 & 47.19 \\ 
& 16.66 & 2 & 4002.48 & 24 & 48.58 & 48.78 & 45.94 & 25.67 & 49.24 & 43.64 & 47.45 \\ 
& 16.73 & 3 & 4643.60 & 25 & 48.98 & 48.90 & 45.94 & 25.67 & 49.24 & 43.74 & 47.71 \\ 
& 19.52 & 1 & 4214.00 & 23 & 51.69 & 52.07 & 45.94 & 25.67 & 49.24 & 44.92 & 50.03 \\ \hline

\parbox[t]{2mm}{\multirow{5}{*}{\rotatebox[origin=c]{90}{\textbf{Revenue}}}} & \textbf{75.86} & 5 & 7010.20 & 27 & 48.87 & 67.38 & 11.78 & 100.00 & 100.00 & 65.61 & 53.79 \\
& 76.19 & 3 & 6782.00 & 26 & 47.86 & 64.69 & 11.78 & 100.00 & 100.00 & 64.86 & 52.55 \\
& 76.20 & 1 & 7087.67 & 28 & 47.83 & 79.18 & 11.78 & 100.00 & 100.00 & 67.76 & 56.08 \\ 
& 77.24 & 4 & \textbf{7226.72} & 28 & 44.71 & 83.31 & 11.78 & 100.00 & 100.00 & 67.96 & 55.31 \\ 
& 78.19 & 2 & 6814.78 & 28 & 51.82 & 83.79 & 11.78 & 25.67 & 100.00 & 54.61 & 56.46 \\ \hline

\end{tabular}
\end{adjustbox}
\caption{Results by run for each fairness index. Unbalanced capacities.}
\label{tab:unbalanced_capacities}
\end{table*}

Table~\ref{tab:comparison_summarized} clearly illustrates the trade-off between revenue maximization and fairness in the balanced scenario: optimizing for fairness leads to a substantially more equitable distribution between RUs, albeit at the cost of reduced fitness and revenue: in the three scenarios tested, fairness in resource allocation improved significantly, albeit at the expense of a substantial reduction in revenue.

When the optimization is driven by revenue maximization, the fitness and revenue values are significantly higher (with a mean fitness of 6295.91 and revenue of 6295.91), but this comes at the expense of much higher inequity, with a mean value of 29.81\% and a larger standard deviation (5.74). In contrast, the fairness-oriented approaches using Jain, Gini, and Atkinson indices yield markedly lower inequity levels; for instance, Jain and Gini optimizations result in mean inequity values of 1.75\% and 2.04\%, respectively, while Atkinson shows a mean inequity of 7.62\% albeit with higher variability.

Furthermore, the fairness-driven runs exhibit more consistent behavior, as evidenced by the lower standard deviations in the fairness metrics and in the importance assigned to each RU. Revenue-driven optimization, while achieving superior economic performance, results in a broader dispersion of the RU allocated importance values, and deviates considerably from the ideal equal distribution. The assigned capacity closely follows the trend observed for total importance, further confirming that the model effectively enforces the balanced capacity constraint. 

\begin{table*}[htbp]
\centering
\begin{adjustbox}{max width=\textwidth}
\begin{tabular}{|c|c|cccc|cccc|cccc|}
\cline{3-14}
\multicolumn{2}{c|}{} & \multicolumn{4}{c|}{\textbf{Balanced}} & \multicolumn{4}{c|}{\textbf{Semi-balanced}} & \multicolumn{4}{c|}{\textbf{Unbalanced}} \\
\cline{3-14}
\multicolumn{2}{c|}{} & \textbf{Jain} & \textbf{Gini} & \textbf{Atkinson} & \textbf{Revenue} & \textbf{Jain} & \textbf{Gini} & \textbf{Atkinson} & \textbf{Revenue} & \textbf{Jain} & \textbf{Gini} & \textbf{Atkinson} & \textbf{Revenue} \\
\hline
\multirow{2}{*}{\textbf{Inequity}} 
   & $\mu$    & \textbf{1.75}    & 2.04    & 7.62    & 29.81   & 5.21    & \textbf{1.93}    & 8.08    & 26.06   & \textbf{16.93}   & 17.13   & 17.21   & 76.74 \\
\cline{2-14}
   & $\sigma$ & 1.48    & \textbf{0.55}    & 8.96    & 5.74    & 4.14    & \textbf{0.82}    & 10.92   & 11.28   & \textbf{0.21}    & 0.54    & 1.30    & 0.96 \\
\hline
\multirow{2}{*}{\textbf{Revenue}}
   & $\mu$    & 2854.10 & 2531.24 & 2389.06 & \textbf{6295.91} & 2571.68 & 3122.67 & 3240.69 & \textbf{6939.10} & 3599.89 & 4296.22 & 4202.07 & \textbf{6984.27} \\
\cline{2-14}
   & $\sigma$ & 487.91  & 372.34  & 662.68  & 443.27  & 1086.35 & 226.29  & 523.83  & 420.45  & 302.95  & 767.78  & 258.44  & 186.94 \\
\hline
\multirow{2}{*}{\textbf{\boldmath $I_1$ (\%)}} 
   & $\mu$    & 35.50   & 36.57   & 39.11   & 33.73   & 49.34   & 49.36   & 51.74   & 56.54   & 49.16   & 49.51   & 49.02   & 48.22 \\
\cline{2-14}
   & $\sigma$ & 5.00    & 4.66    & 6.45    & 12.18   & 5.78    & 1.09    & 4.51    & 10.39   & 0.47    & 0.86    & 1.57    & 2.55 \\
\hline
\multirow{2}{*}{\textbf{\boldmath $I_2$ (\%)}} 
   & $\mu$    & 36.54   & 36.17   & 41.34   & 44.72   & 49.26   & 48.65   & 51.29   & 57.61   & 49.16   & 48.96   & 49.34   & 75.67 \\
\cline{2-14}
   & $\sigma$ & 4.81    & 4.38    & 5.71    & 14.75   & 9.24    & 1.92    & 3.57    & 4.67    & 0.27    & 0.81    & 1.54    & 9.03 \\
\hline
\multirow{2}{*}{\textbf{\boldmath $I_3$ (\%)}} 
   & $\mu$    & 37.02   & 36.43   & 40.51   & 56.55   & 50.39   & 49.12   & 44.47   & 45.64   & 45.94   & 45.94   & 45.94   & 11.78 \\
\cline{2-14}
   & $\sigma$ & 4.79    & 5.10    & 6.64    & 11.97   & 8.21    & 1.28    & 13.33   & 15.42   & 0.00    & 0.00    & 0.00    & 0.00 \\
\hline
\multirow{2}{*}{\textbf{\boldmath $I_4$ (\%)}} 
   & $\mu$    & 37.02   & 36.61   & 39.96   & 55.56   & 50.63   & 49.30   & 50.91   & 70.71   & 25.67   & 25.67   & 25.67   & 85.13 \\
\cline{2-14}
   & $\sigma$ & 4.12    & 4.04    & 6.08    & 9.84    & 8.25    & 1.95    & 4.99    & 2.74    & 0.00    & 0.00    & 0.00    & 33.24 \\
\hline
\multirow{2}{*}{\textbf{\boldmath $I_5$ (\%)}} 
   & $\mu$    & 37.24   & 36.44   & 33.59   & 61.07   & 46.73   & 48.04   & 50.43   & 58.03   & 49.24   & 49.24   & 49.24   & 100.00 \\
\cline{2-14}
   & $\sigma$ & 4.65    & 3.19    & 8.83    & 4.38    & 13.92   & 0.63    & 5.87    & 16.05   & 0.00    & 0.00    & 0.00    & 0.00 \\
\hline
\multirow{2}{*}{\makecell{\textbf{Total} \\ \textbf{Importance}}} 
   & $\mu$    & 36.66   & 36.44   & 38.90   & 50.33   & 49.27   & 48.89   & 49.76   & 57.71   & 43.83   & 43.86   & 43.84   & 64.16 \\
\cline{2-14}
   & $\sigma$ & 4.56    & 4.19    & 3.69    & 3.50    & 8.87    & 1.13    & 1.60    & 2.45    & 0.13    & 0.32    & 0.62    & 5.50 \\
\hline
\multirow{2}{*}{\makecell{\textbf{Assigned} \\ \textbf{capacity}}} 
   & $\mu$    & 36.67   & 36.44   & 38.90   & 50.33   & 49.47   & 49.00   & 49.91   & 57.03   & 47.87   & 48.02   & 47.84   & 54.84 \\
\cline{2-14}
   & $\sigma$ & 4.56    & 4.18    & 3.69    & 3.50    & 8.14    & 1.24    & 1.68    & 1.42    & 0.30    & 0.67    & 1.27    & 1.64 \\
\hline
\end{tabular}
\end{adjustbox}
\caption{Results by run for each fairness index in the three different framework capacity scenarios: Balanced, Semi-balanced and Unbalanced.}
\label{tab:comparison_summarized}
\end{table*}

Table \ref{tab:comparison_summarized} also summarizes a clear trade-off between fairness and revenue across the three framework capacity scenarios. In all cases, fairness-guided methods using indices such as Jain, Gini, and Atkinson significantly reduce inequity compared to the revenue maximization approach. For instance, in the balanced scenario, the Jain method achieves an inequity of just 1.75, while the revenue-based approach yields an inequity of 29.81. However, these improvements in fairness come at the cost of revenue, with fairness-oriented methods reporting revenues as low as 2854.10, in contrast to 6295.91 achieved by the revenue maximization strategy. Moreover, inequity worsens in the strongly unbalanced scenario, reflecting the increased challenge of attaining an equitable distribution when capacities are highly disparate. Similar trends are observed in the semi-balanced and unbalanced cases, thereby highlighting the inherent compromise between achieving equitable resource distribution and maintaining high revenue performance.

\begin{figure}[!ht]
    \centering
    \begin{subfigure}[b]{0.32\textwidth}
        \centering
        \includegraphics[width=\textwidth]{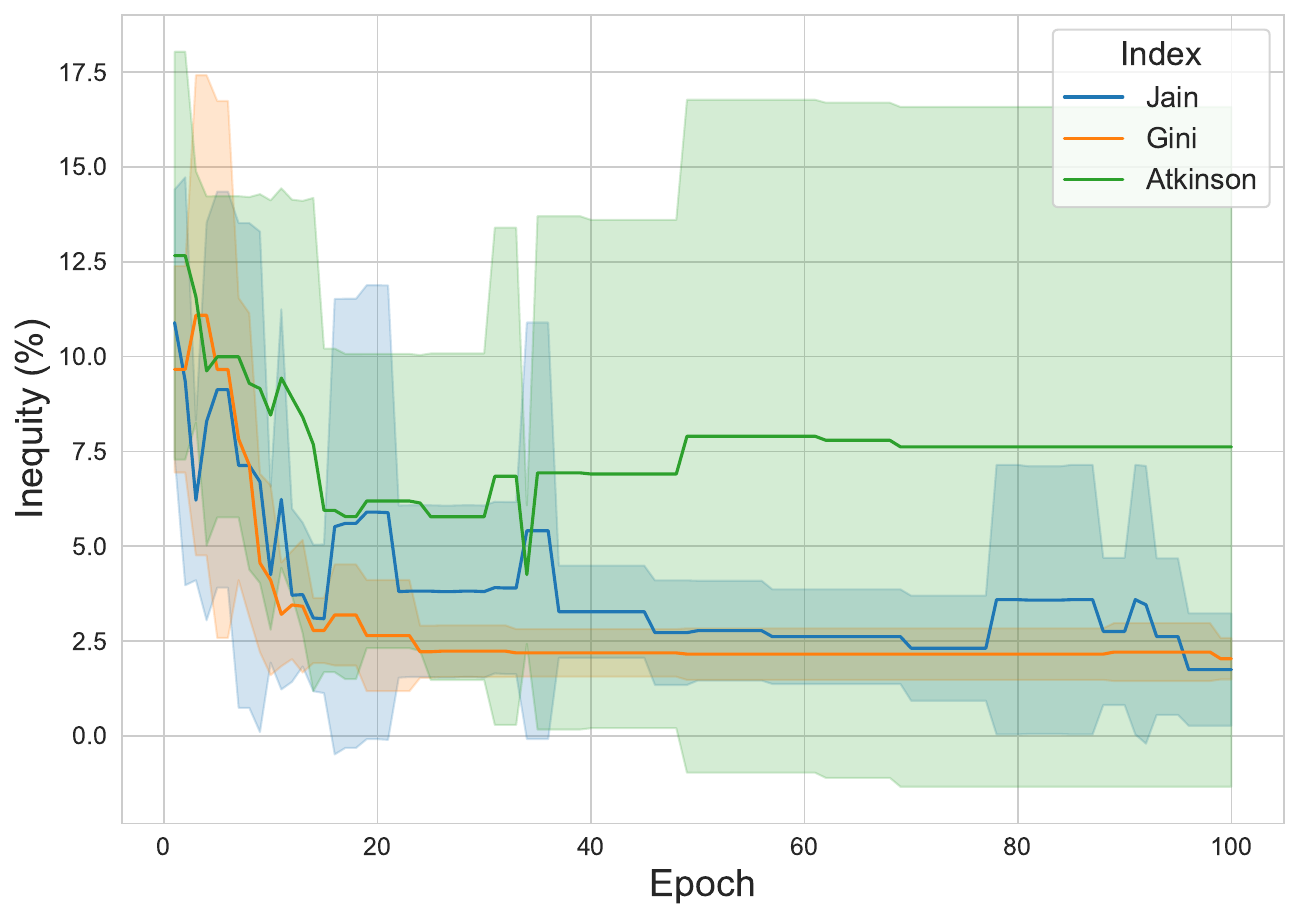}
        \caption{Balanced}
        \label{fig:inequity_balanced_evo}
    \end{subfigure}
    \hfill
    \begin{subfigure}[b]{0.32\textwidth}
        \centering
        \includegraphics[width=\textwidth]{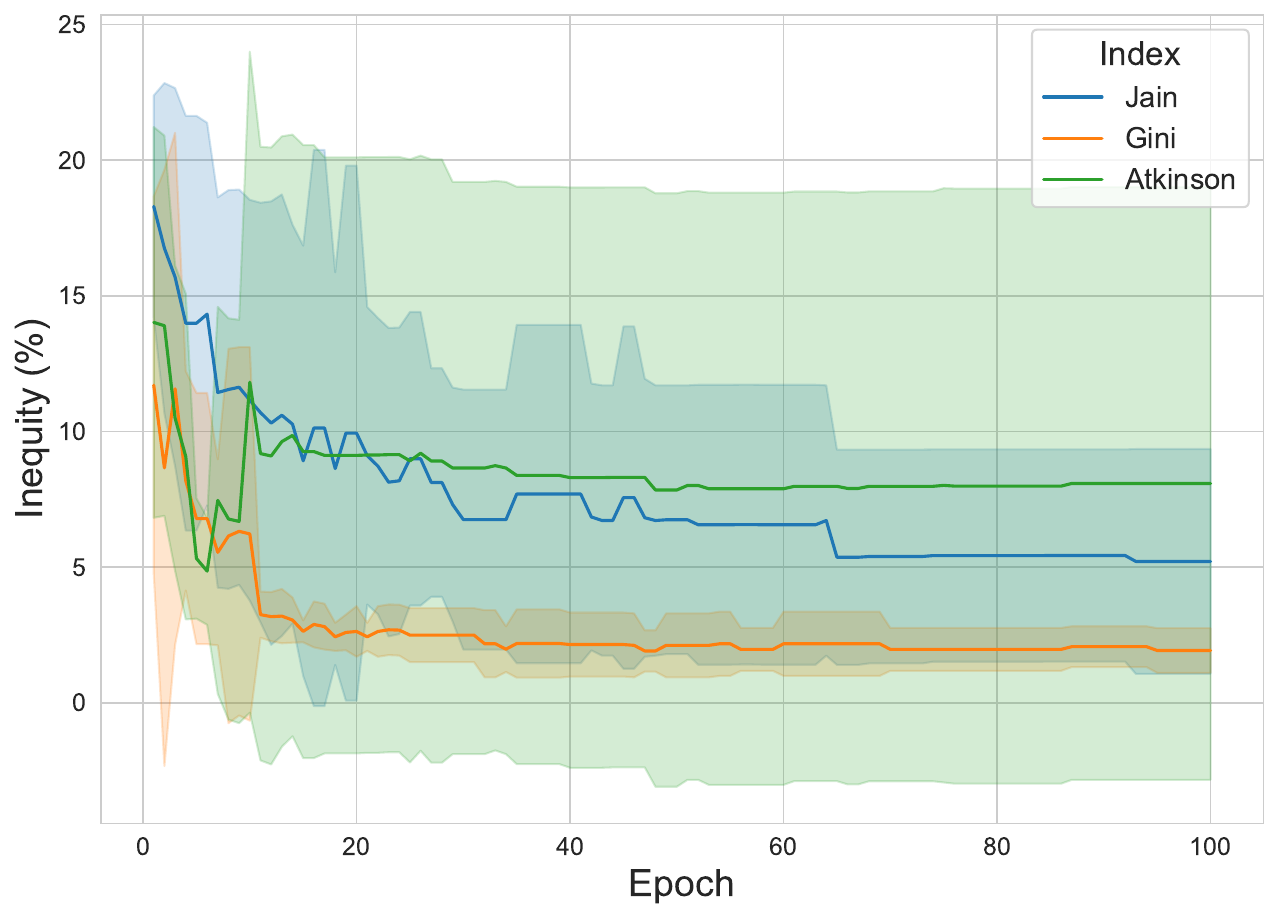}
        \caption{Semi-balanced}
        \label{fig:inequity_semi-balanced_evo}
    \end{subfigure}
    \hfill
    \begin{subfigure}[b]{0.32\textwidth}
        \centering
        \includegraphics[width=\textwidth]{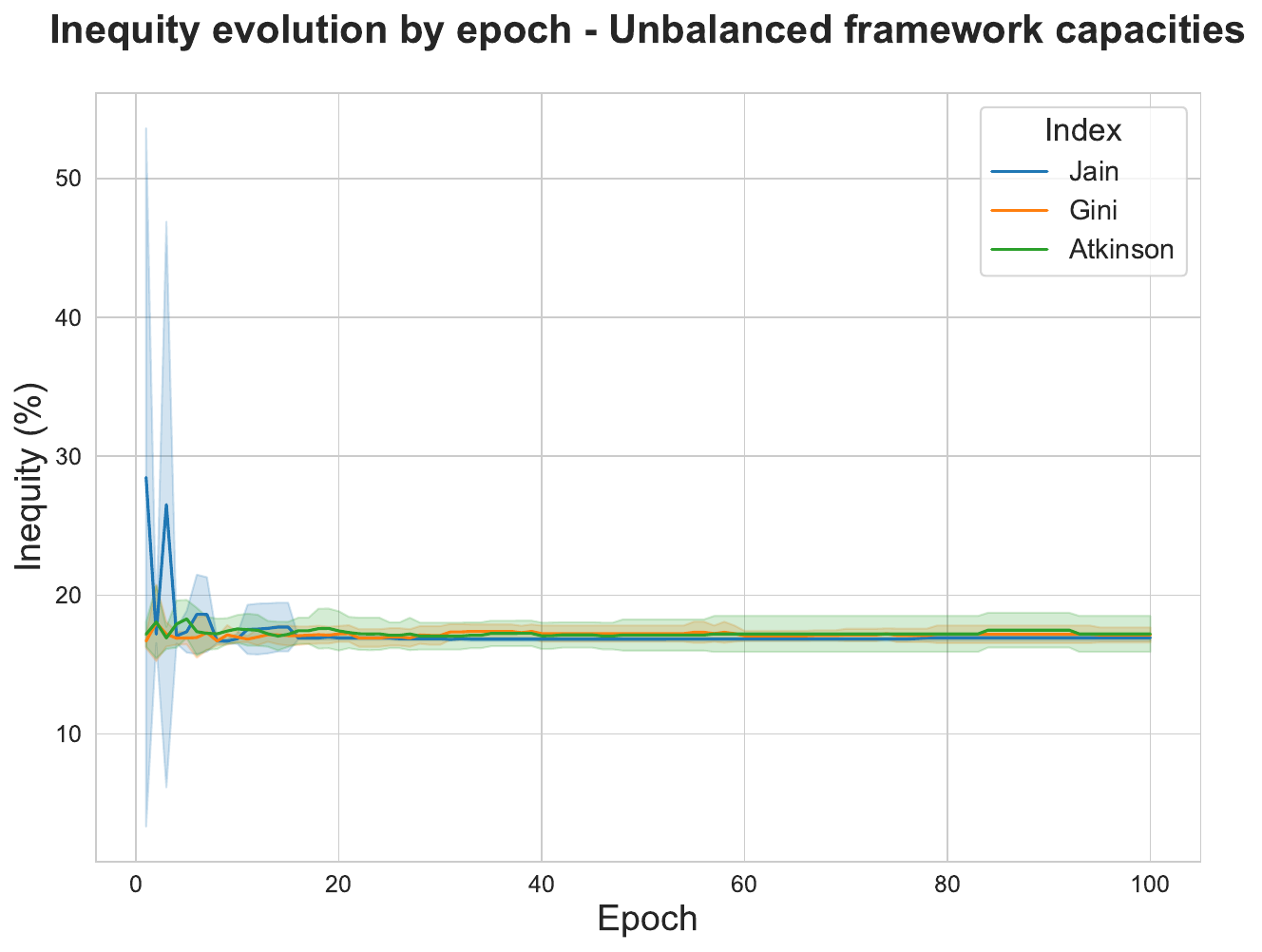}
        \caption{Unbalanced}
        \label{fig:inequity_unbalanced_evo}
    \end{subfigure}
    \caption{Evolution of inequity values for the three fairness indices (Jain, Gini, and Atkinson) over training epochs.}
    \label{fig:inequity_comparison}
\end{figure}

Figure~\ref{fig:inequity_comparison} presents the evolution of inequity (expressed as a percentage) over training epochs. In all three cases, it can be observed that fairness-driven approaches tend to reduce inequity levels throughout the training process. Consistent with the data shown in Table~\ref{tab:comparison_summarized}, the Jain approach achieves the lowest inequity, while the lowest standard deviation between runs is obtained with the Gini approach. On the other hand, both the mean inequity and the standard deviation are higher with the Atkinson coefficient, as shown in Figure~\ref{fig:inequity_balanced_evo}. In the semi-balanced scenario (Figure~\ref{fig:inequity_semi-balanced_evo}), the Gini approach attains the lowest inequity and standard deviation, followed by Jain and Atkinson. Finally, in the unbalanced case (Figure~\ref{fig:inequity_unbalanced_evo}), the Jain approach yields the best results, although the differences between the fairness indices are considerably smaller in this scenario.

\begin{figure}[!ht]
    \centering
    \begin{subfigure}[b]{0.32\textwidth}
        \centering
        \includegraphics[width=\textwidth]{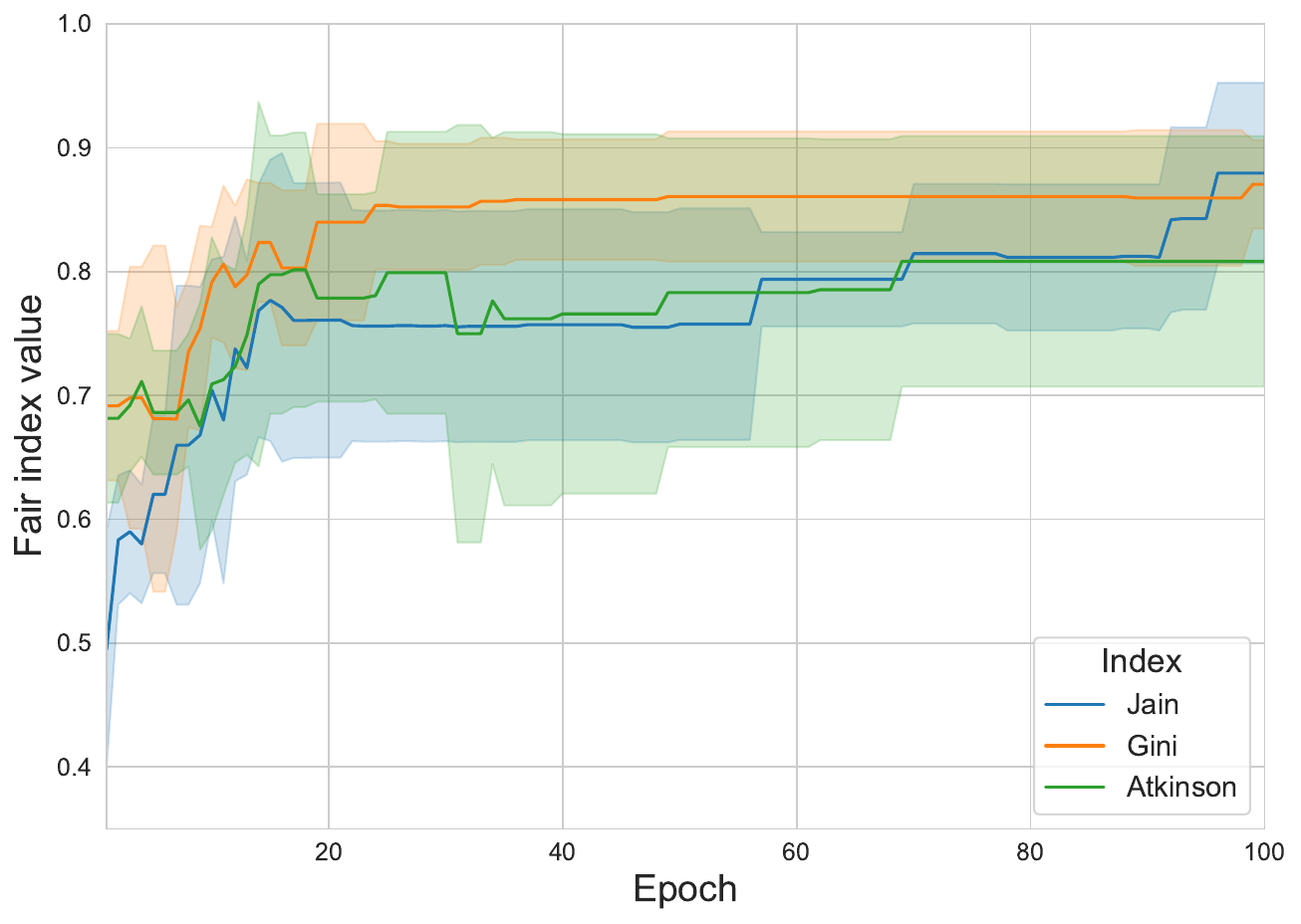}
        \caption{Balanced framework.}
        \label{fig:balanced_fair_index_evo}
    \end{subfigure}
    \hfill
    \begin{subfigure}[b]{0.32\textwidth}
        \centering
        \includegraphics[width=\textwidth]{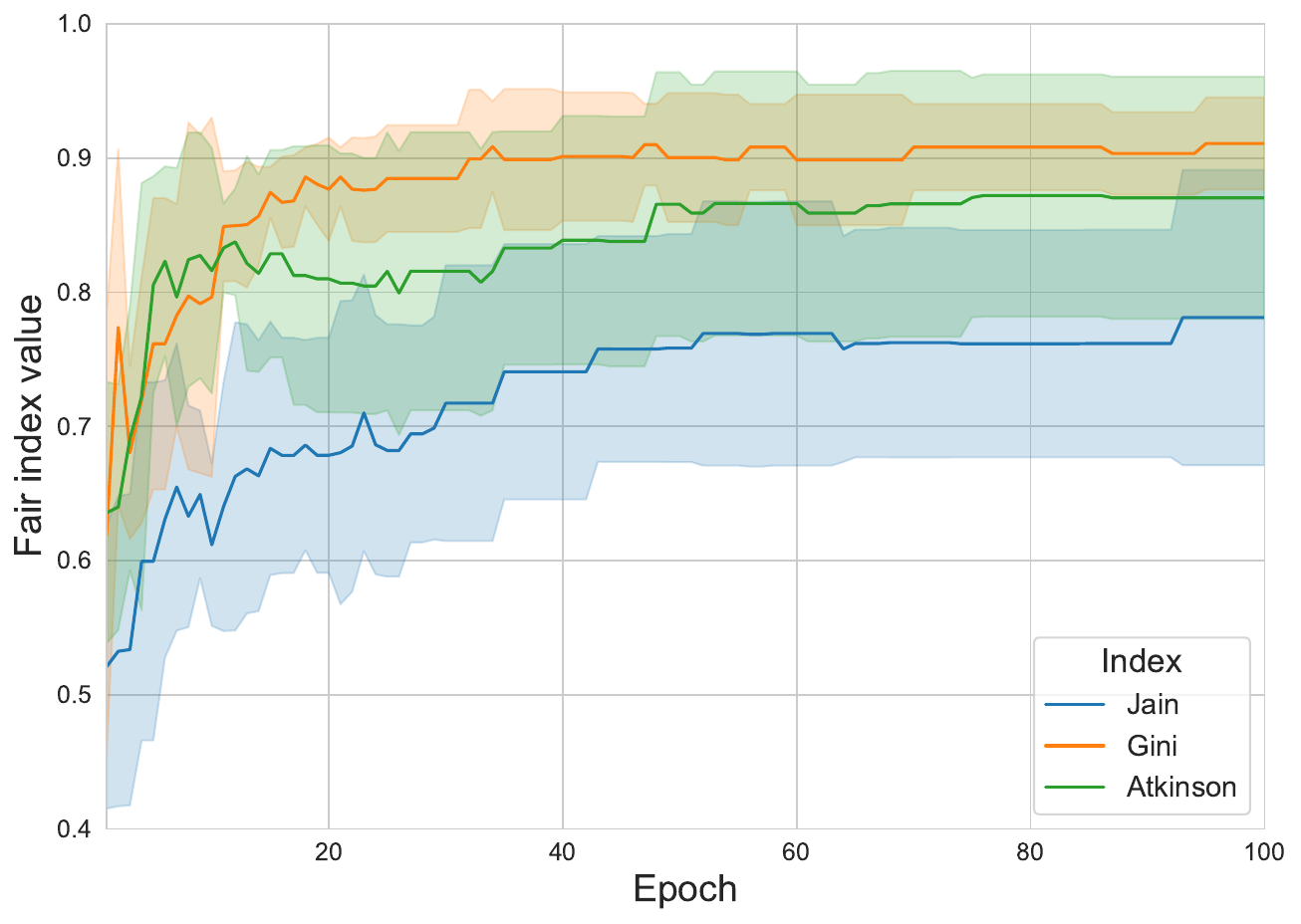}
        \caption{Semi-balanced framework.}
        \label{fig:semibalanced_fair_index_evo}
    \end{subfigure}
    \hfill
    \begin{subfigure}[b]{0.32\textwidth}
        \centering
        \includegraphics[width=\textwidth]{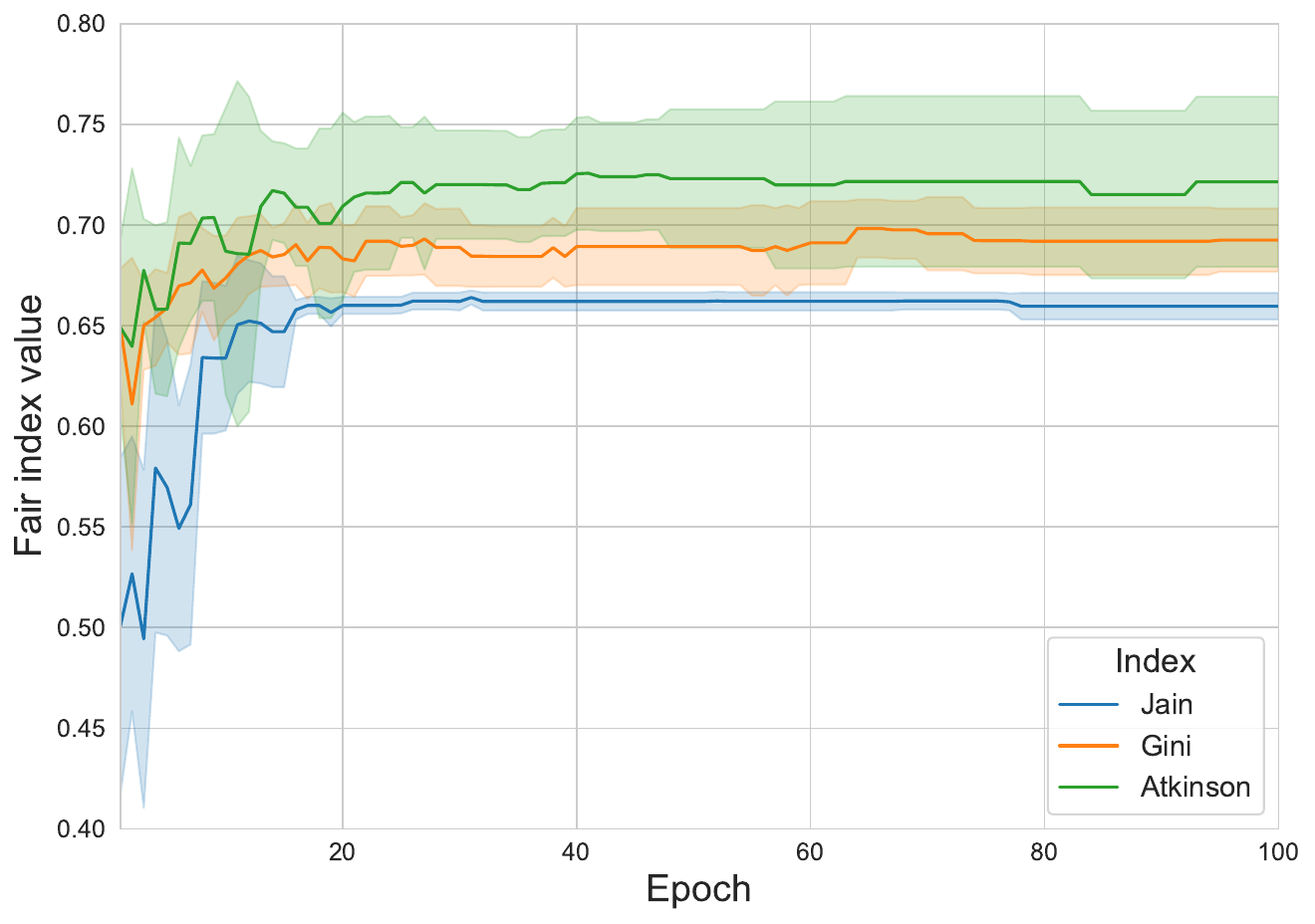}
        \caption{Unbalanced framework.}
        \label{fig:unbalanced_fair_index_evo}
    \end{subfigure}
    
    \caption{Evolution of fairness index values for the three fairness indices (Jain, Gini, and Atkinson) over training epochs.}
    \label{fig:comparison_fair_index_evo}
\end{figure}

Figure~\ref{fig:comparison_fair_index_evo} shows the evolution of each fairness metric over training epochs. Tracking the Jain, Gini, and Atkinson indices reveals a consistent trend towards higher equity (i.e., values closer to 1). In all three cases (Figures~\ref{fig:balanced_fair_index_evo}, \ref{fig:semibalanced_fair_index_evo}, and \ref{fig:unbalanced_fair_index_evo}), the Jain approach exhibits the slowest convergence; however, it ultimately returns the best results in the balanced scenario (Figure~\ref{fig:balanced_fair_index_evo}). The Atkinson and Gini indices display similar convergence behaviors, with the Gini index achieving the best results in the semi-balanced case (Figure~\ref{fig:semibalanced_fair_index_evo}) and the Atkinson index excelling in the unbalanced scenario (Figure~\ref{fig:unbalanced_fair_index_evo}). However, it is important to note that the Atkinson approach consistently exhibits a higher standard deviation in all cases.

\begin{figure}[!ht]
    \centering
    \begin{subfigure}[b]{0.32\textwidth}
        \centering
        \includegraphics[width=\textwidth]{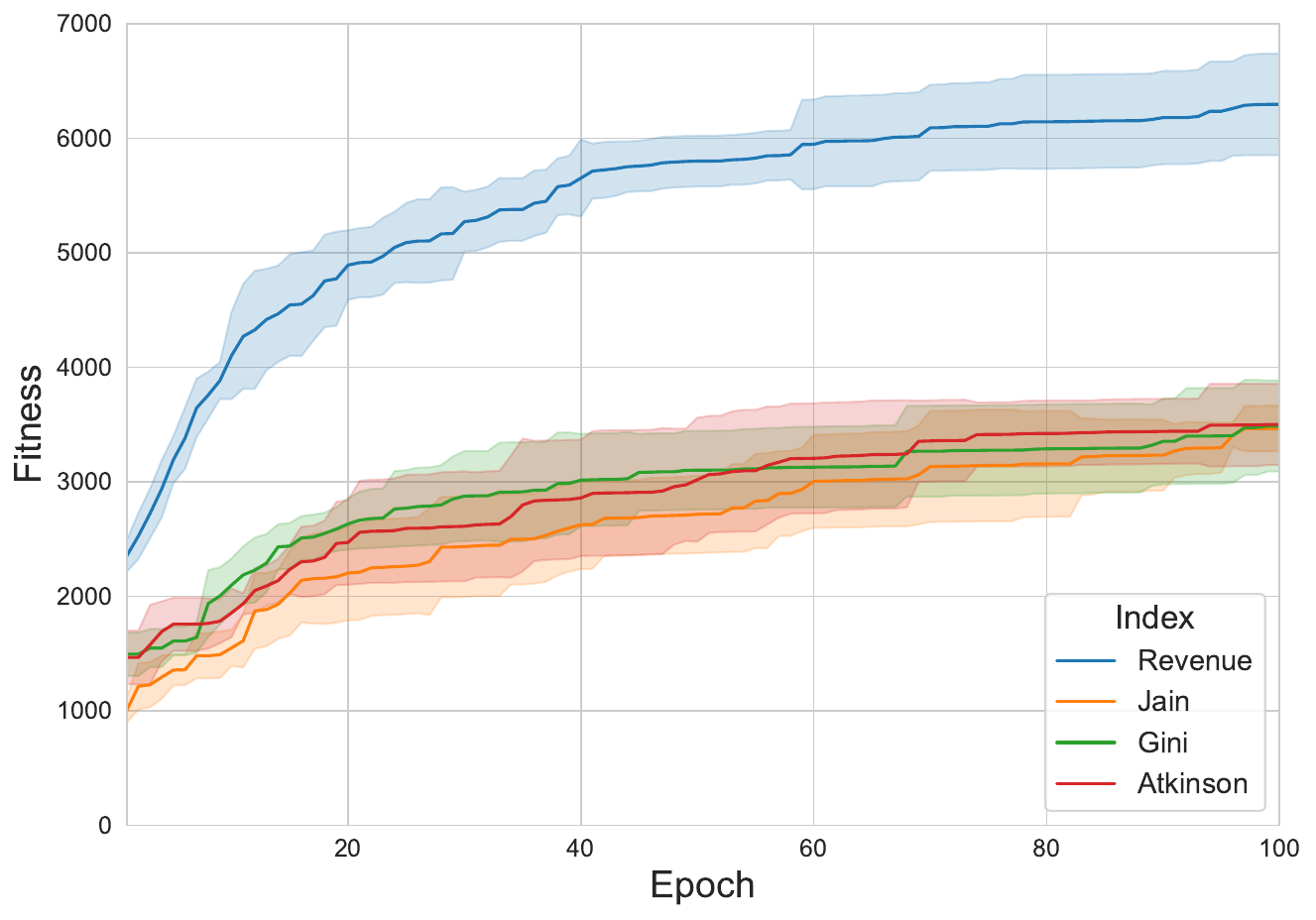}
        \caption{Fitness evolution over training epochs.}
        \label{fig:fitness_balanced_evo}
    \end{subfigure}
    \hfill
    \begin{subfigure}[b]{0.32\textwidth}
       \centering
        \includegraphics[width=\textwidth]{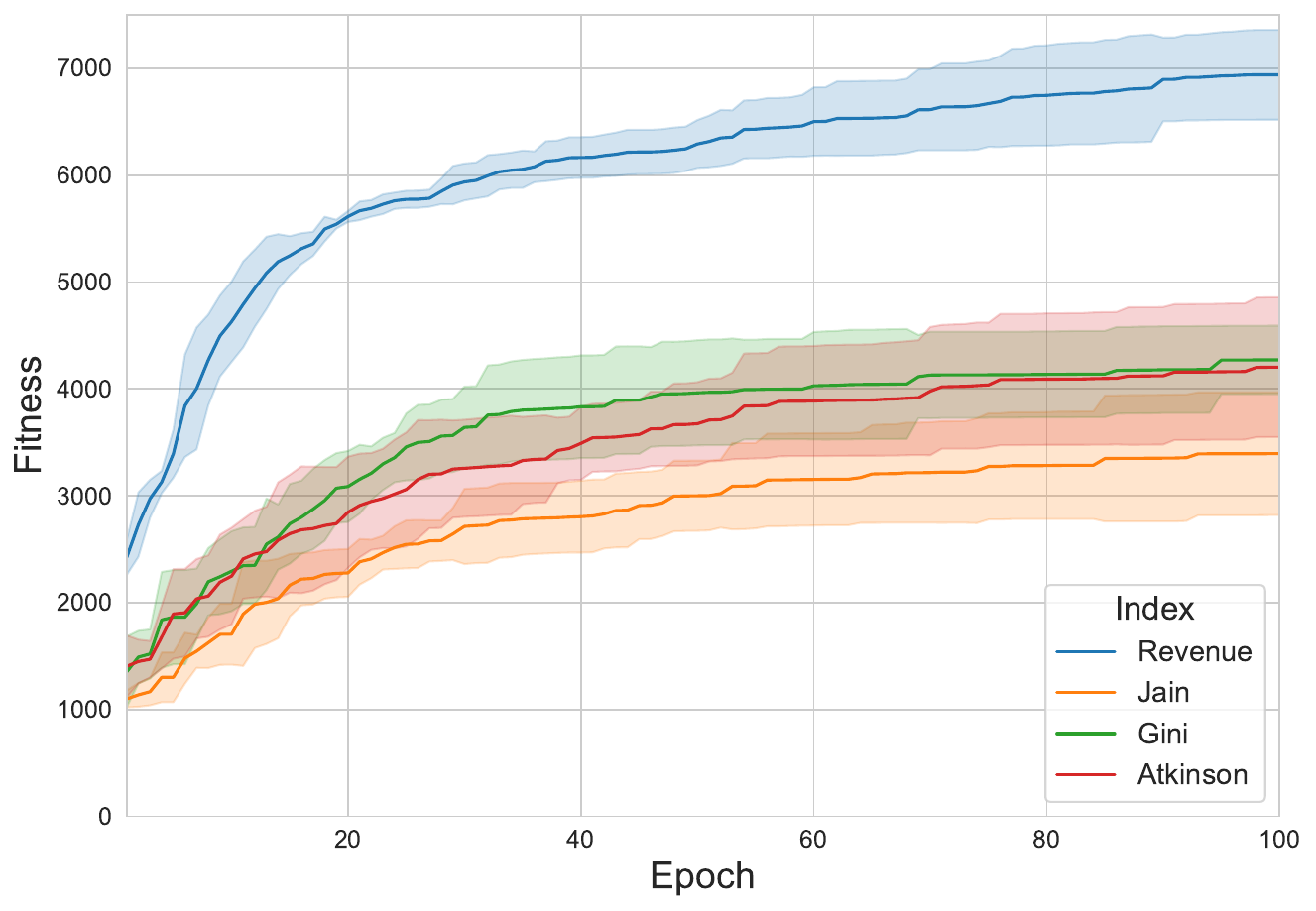}
        \caption{Fitness evolution over training epochs.}
        \label{fig:fitness_semibalanced_evo}
    \end{subfigure}
    \hfill    
    \begin{subfigure}[b]{0.32\textwidth}
        \centering
        \includegraphics[width=\textwidth]{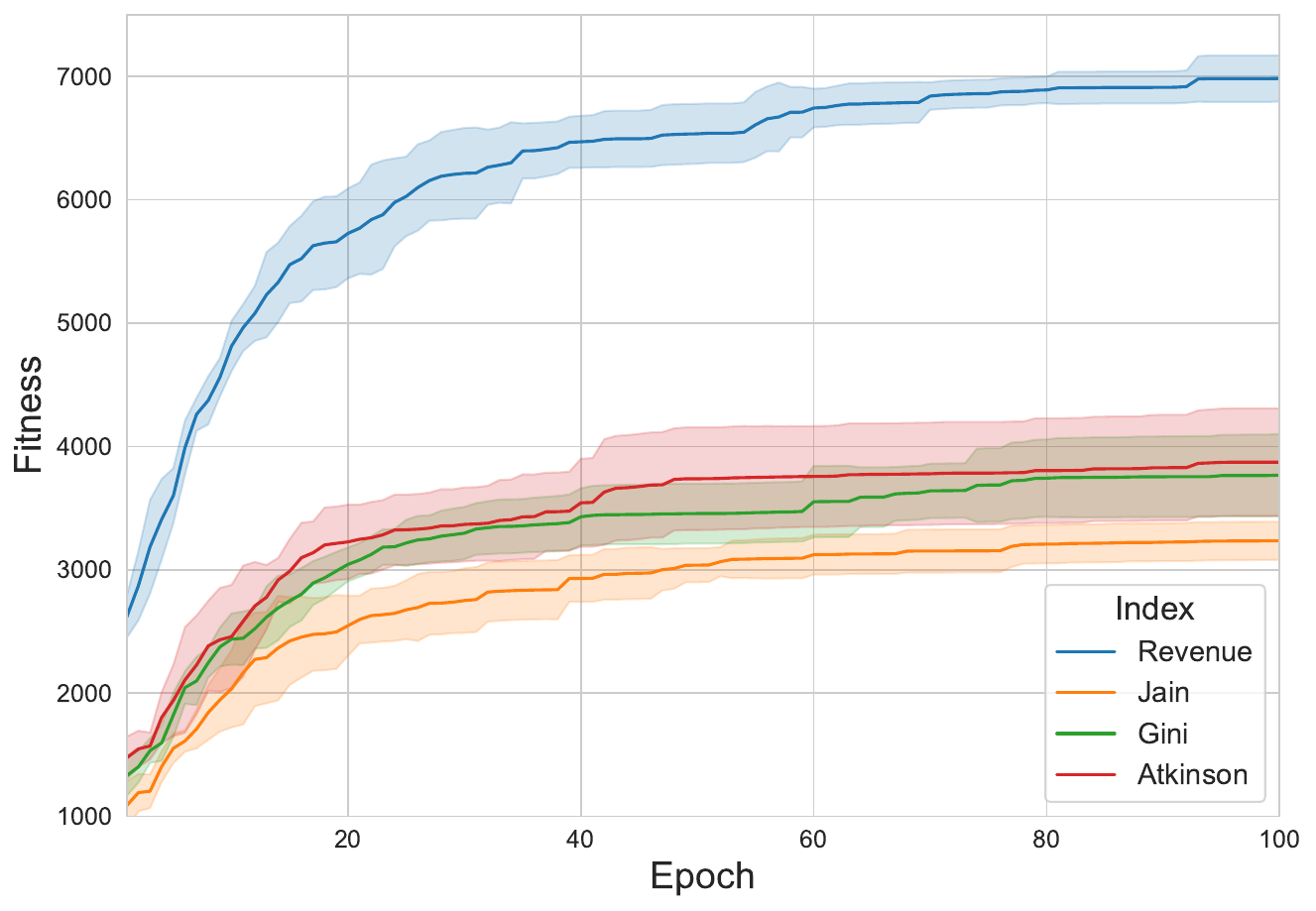}
        \caption{Fitness evolution over training epochs.}
        \label{fig:fitness_unbalanced_evo}
    \end{subfigure}
    
    \caption{Fitness evolution for the three fairness indices (Jain, Gini, and Atkinson) over training epochs.}
    \label{fig:comparison_fitness_index_evo}
\end{figure}

Figure~\ref{fig:comparison_fitness_index_evo} illustrates the evolution of the fitness value over training epochs. The revenue maximization approach attains higher fitness values, whereas the fairness-oriented methods yield lower values, as expected from Equation~\ref{eq:objective_function}, since revenue is multiplied by a factor in the range $[0,1]$. Although the fairness-driven runs achieve lower fitness values, they still converge steadily, demonstrating that the approach maintains acceptable performance while enforcing equitable allocation. This trade-off underscores the inherent challenge of balancing economic efficiency with fairness.

\begin{figure}[!ht]
    \centering
    \begin{subfigure}[b]{0.32\textwidth}
        \centering
        \includegraphics[width=\textwidth]{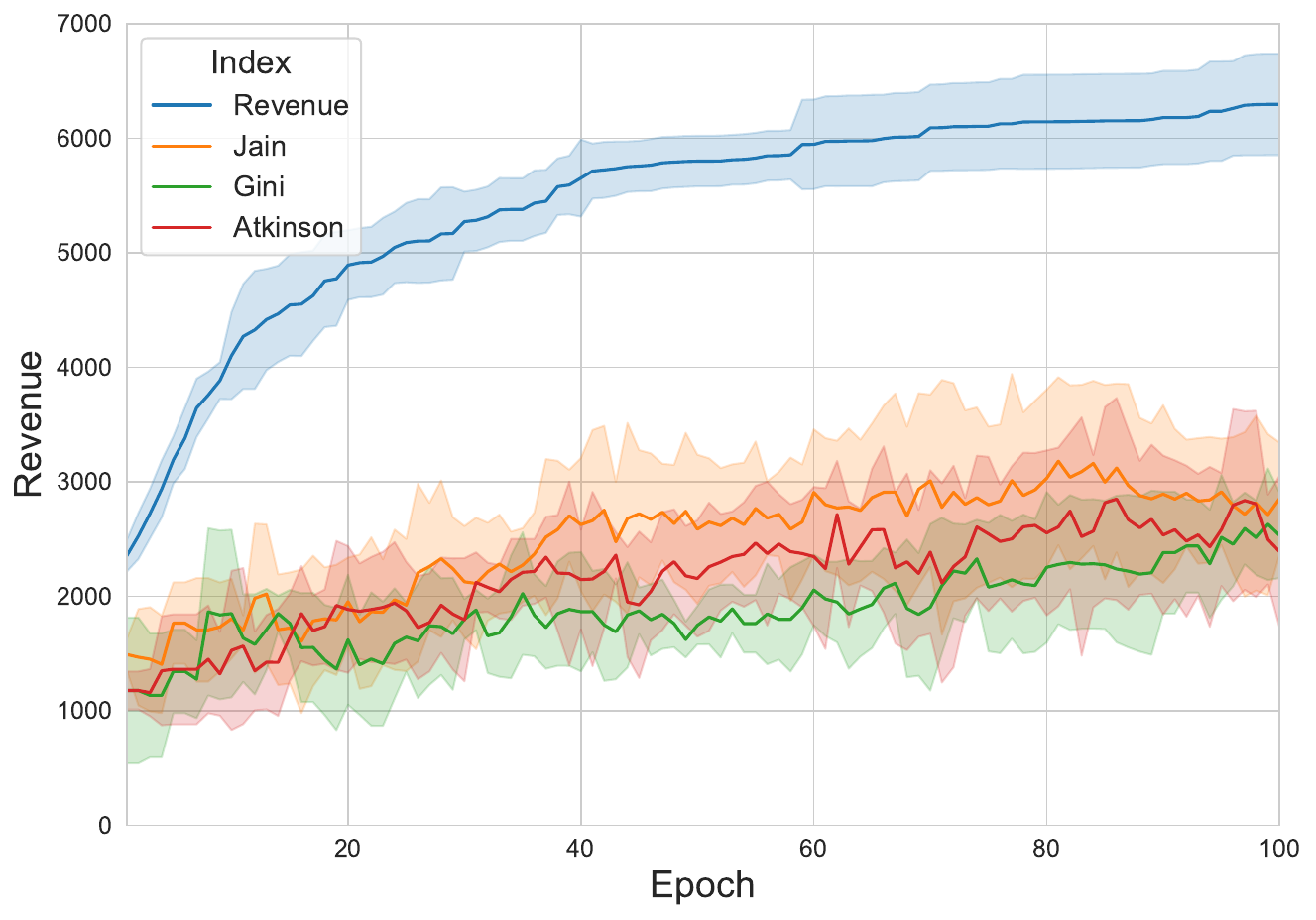}
        \caption{Revenue evolution over training epochs.}
        \label{fig:revenue_balanced_evo}
    \end{subfigure}
    \hfill
    \begin{subfigure}[b]{0.32\textwidth}
       \centering
        \includegraphics[width=\textwidth]{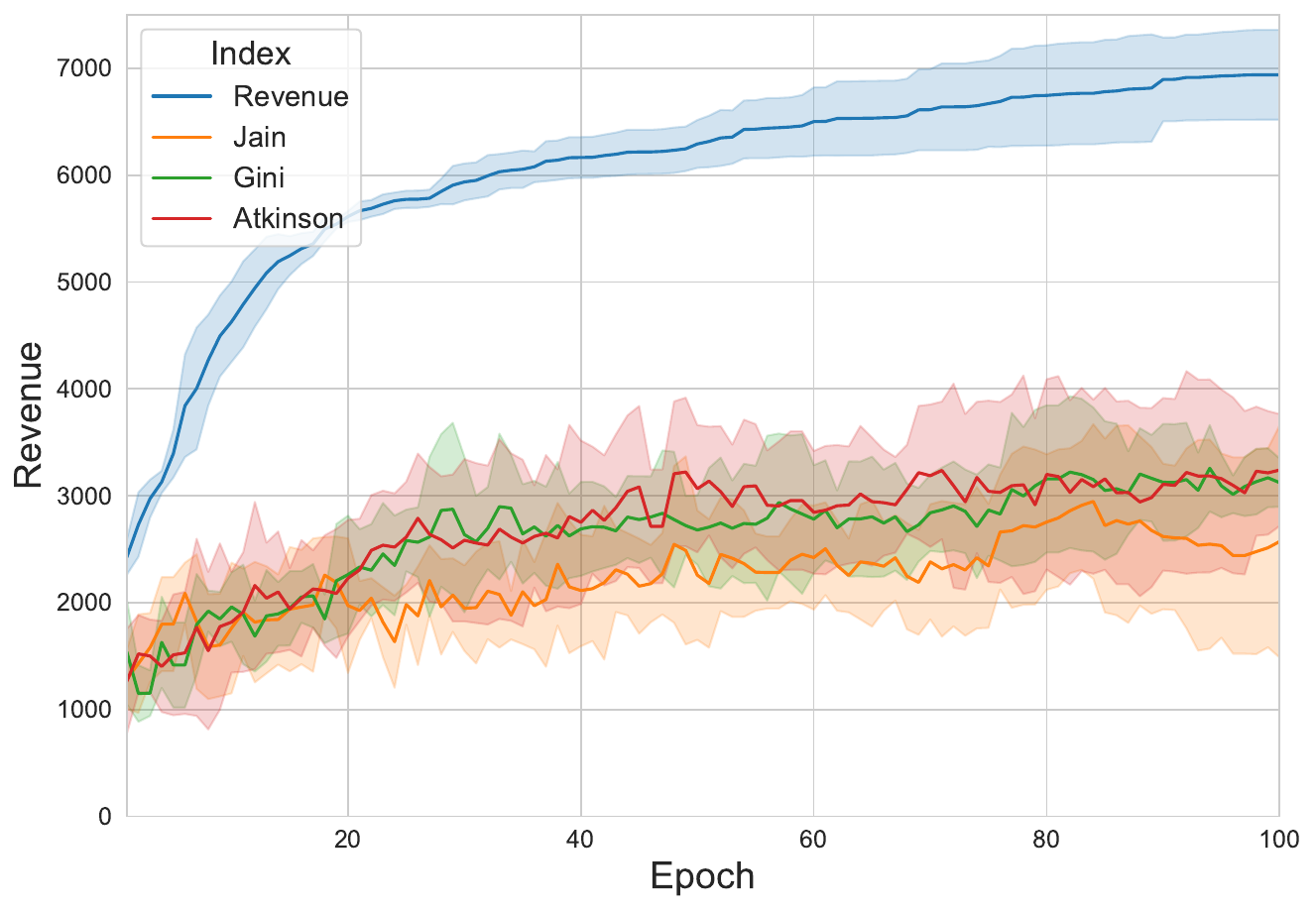}
        \caption{Revenue evolution over training epochs.}
        \label{fig:revenue_semibalanced_evo}
    \end{subfigure}
    \hfill
    \begin{subfigure}[b]{0.32\textwidth}
        \centering
        \includegraphics[width=\textwidth]{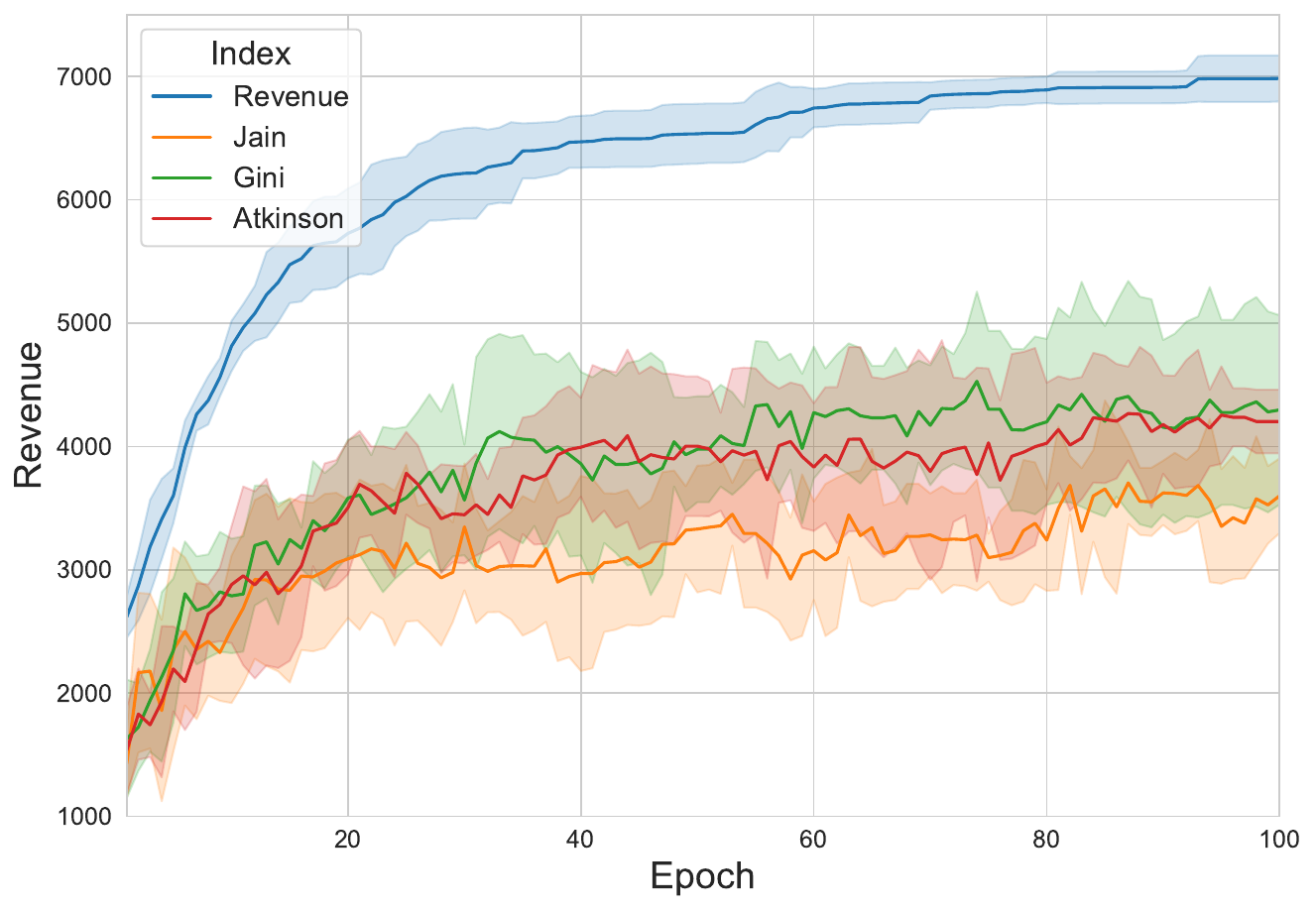}
        \caption{Revenue evolution over training epochs.}
        \label{fig:revenue_unbalanced_evo}
    \end{subfigure}
    
    \caption{Revenue evolution for the three fairness indices (Jain, Gini, and Atkinson) over training epochs.}
    \label{fig:comparison_revenue_index_evo}
\end{figure}

Figure~\ref{fig:comparison_revenue_index_evo} illustrates the evolution of revenue over training epochs. Revenue-based optimization consistently achieves higher revenue values compared to fairness-oriented approaches (Jain, Gini, and Atkinson), highlighting the inherent trade-off between maximizing revenue and ensuring equitable service allocation. The smooth and sustained convergence of the revenue curve is attributed to the fact that its objective function is specifically profit maximization. In contrast, for equity-focused cases— where the objective function is defined differently (see Equation~\ref{eq:objective_function})— greater variability is observed in the curves.

The results presented in this section illustrate the trade-off between economic efficiency and fairness. Fairness-driven methods using the Jain, Gini, and Atkinson indices consistently achieve lower inequity levels and more balanced resource allocations across the balanced, semi-balanced, and unbalanced scenarios. In this regard, it is worth noting that in the case where framework capacities are highly unbalanced, it is particularly difficult to obtain an equitable allocation, which poses a challenge in solving the infrastructure allocation problem in this scenario. However, these approaches incur a penalty in revenue. 

\section{Conclusions and future works}\label{sec:Conclusions}

This paper presents a novel approach to equitable railway timetabling in liberalized markets, integrating a genetic algorithm with adapted fairness metrics. By formulating a mathematical model that represents the service requests of \glspl{RU} and by incorporating service importance, it has been able to adjust traditional equity indices (Jain, Gini and Atkinson) to the context of railway timetabling. The introduction of a sensitivity parameter, $\alpha$, in these metrics proved to be useful in amplifying differences between fair and unfair allocations for a more accurate assessment of equity. Experimental results showed that while revenue-driven strategies yielded higher economic gains, they also produced significantly skewed distributions between \glspl{RU}. In contrast, the fairness-oriented approach consistently achieved more balanced resource allocations at the expense of reduced revenue, thereby highlighting the inherent trade-off between efficiency and equity in railway scheduling.

The proposed methodology also underscores the importance of calibrating optimization objectives based on the specific context of the system. Results showed that setting $\alpha=25$ for the Jain and Atkinson indices, and $\alpha=10$ for the Gini index, leads to an optimal enhancement of fairness sensitivity when compared to the baseline revenue maximization. This calibration not only improved the fairness indices but also ensured that slight deviations from an equitable allocation were properly penalized. Despite these promising outcomes, the proposed approach still faces limitations, especially when addressing scenarios characterized by highly unbalanced infrastructure capacities between the \glspl{RU}. The challenge lies in reconciling the dual objectives of maximizing revenue and ensuring fair access to infrastructure, a problem that intensifies as disparities in RU capacities widen.

Future work could explore alternative or composite fairness indicators. Finally, validating the proposed methodology using real operational data from \glspl{IM} and \glspl{RU} would provide valuable insights into the practical applicability of the approach and help refine the model parameters to better reflect real-world constraints.

\section*{CRediT authorship contribution statement}
\textbf{David Muñoz-Valero:} Writing – original draft, Visualization, Validation, Software, Methodology, Data curation, Conceptualization. \textbf{Juan Moreno-Garcia:} Writing – original draft, Validation, Resources, Supervision, Methodology, Formal analysis, Conceptualization, Funding acquisition. \textbf{Julio Alberto López-Gómez:} Writing – review \& editing, Supervision, Methodology, Conceptualization, Formal analysis. \textbf{Enrique Adrian Villarrubia-Martin:} Writing – review \& editing, Data curation, Software.

\section*{Declaration of competing interest}
The authors declare that they have no known competing financial interests or personal relationships that could have appeared to influence the work reported in this paper.

\section*{Data availability}
The software is publicly available in the following repository: \url{https://github.com/DavidMunozValero/GSA_M}.




\section*{Acknowledgments}

This work was supported by grants PID2020-112967GB-C32 and PID2020-112967GB-C33 funded by MCIN/AEI/10.13039/501100011033, by ERDF A Way of Making Europe and the Research Vice-Rectory of the Universidad de Castilla-La Mancha. It was completed when Enrique Adrian Villarrubia-Martin was a predoctoral fellow at the Universidad de Castilla-La Mancha funded by the European Social Fund Plus (ESF+).

\bibliographystyle{elsarticle-num-names}
\bibliography{references.bib}

\begin{thebibliography}{30}
\expandafter\ifx\csname natexlab\endcsname\relax\def\natexlab#1{#1}\fi
\providecommand{\url}[1]{\texttt{#1}}
\providecommand{\href}[2]{#2}
\providecommand{\path}[1]{#1}
\providecommand{\DOIprefix}{doi:}
\providecommand{\ArXivprefix}{arXiv:}
\providecommand{\URLprefix}{URL: }
\providecommand{\Pubmedprefix}{pmid:}
\providecommand{\doi}[1]{\href{http://dx.doi.org/#1}{\path{#1}}}
\providecommand{\Pubmed}[1]{\href{pmid:#1}{\path{#1}}}
\providecommand{\bibinfo}[2]{#2}
\ifx\xfnm\relax \def\xfnm[#1]{\unskip,\space#1}\fi
\bibitem[{Kallas(2011)}]{COM2011}
\bibinfo{author}{S.~Kallas}, \bibinfo{title}{{White Paper on Transport: Roadmap to a Single European Transport Area--towards a Competitive and Resource-efficient Transport System}}, \bibinfo{publisher}{Office for Official Publications of the European Communities}, \bibinfo{year}{2011}.
\bibitem[{Ait~Ali and Eliasson(2022)}]{Ait2021}
\bibinfo{author}{A.~Ait~Ali}, \bibinfo{author}{J.~Eliasson},
\newblock \bibinfo{title}{{European railway deregulation: an overview of market organization and capacity allocation}},
\newblock \bibinfo{journal}{Transportmetrica A: Transport Science} \bibinfo{volume}{18} (\bibinfo{year}{2022}) \bibinfo{pages}{594--618}. \DOIprefix\doi{10.1080/23249935.2021.1885521}.
\bibitem[{Yao et~al.(2024)Yao, Li, Mo, D'Ariano, and Appolloni}]{Yao2024Bi-objectiveLine}
\bibinfo{author}{Y.~Yao}, \bibinfo{author}{P.~Li}, \bibinfo{author}{P.~Mo}, \bibinfo{author}{A.~D'Ariano}, \bibinfo{author}{A.~Appolloni},
\newblock \bibinfo{title}{{Bi-objective optimization of timetable and rolling stock schedule for an urban rail passenger and freight line}},
\newblock \bibinfo{journal}{Computers {\&} Industrial Engineering} \bibinfo{volume}{194} (\bibinfo{year}{2024}) \bibinfo{pages}{110394}. \URLprefix \url{https://www.sciencedirect.com/science/article/pii/S0360835224005151}. \DOIprefix\doi{10.1016/J.CIE.2024.110394}.
\bibitem[{{Maria Teresa Pena-Alcaraz}(2015)}]{Pena2015}
\bibinfo{author}{{Maria Teresa Pena-Alcaraz}}, \bibinfo{title}{{Analysis of Capacity Pricing and Allocation Mechanisms in Shared Railway Systems}}, \bibinfo{type}{Technical Report}, \bibinfo{year}{2015}.
\bibitem[{Be{\v{s}}inovi{\'{c}} et~al.(2024)Be{\v{s}}inovi{\'{c}}, Garc{\'{i}}a-R{\'{o}}denas, L{\'{o}}pez-Garc{\'{i}}a, L{\'{o}}pez-G{\'{o}}mez, and Mart{\'{i}}n-Baos}]{Besinovic2024}
\bibinfo{author}{N.~Be{\v{s}}inovi{\'{c}}}, \bibinfo{author}{R.~Garc{\'{i}}a-R{\'{o}}denas}, \bibinfo{author}{M.~L. L{\'{o}}pez-Garc{\'{i}}a}, \bibinfo{author}{J.~A. L{\'{o}}pez-G{\'{o}}mez}, \bibinfo{author}{J.~{\'{A}}. Mart{\'{i}}n-Baos},
\newblock \bibinfo{title}{{The time slot allocation problem in liberalised passenger railway markets: a multi-objective approach}},
\newblock \bibinfo{journal}{arXiv preprint arXiv:2401.12073}  (\bibinfo{year}{2024}).
\bibitem[{{\'{A}}lvarez(2017)}]{Caramello2017}
\bibinfo{author}{M.~{\'{A}}.~C. {\'{A}}lvarez}, \bibinfo{title}{{Public service rail transport in the European Union: an overview}}, \bibinfo{publisher}{CER}, \bibinfo{year}{2017}.
\bibitem[{Smoliner et~al.(2018)Smoliner, Walter, and Marschnig}]{Smoliner2018}
\bibinfo{author}{M.~S. Smoliner}, \bibinfo{author}{S.~Walter}, \bibinfo{author}{S.~Marschnig},
\newblock \bibinfo{title}{{Optimal Coordination of Timetable and Infrastructure Development in a Liberalised Railway Market}},
\newblock \bibinfo{journal}{Journal of Management and Financial Sciences}  (\bibinfo{year}{2018}) \bibinfo{pages}{97--115}.
\bibitem[{Caimi et~al.(2018)Caimi, Fischer, and Schlechte}]{Schlechte2012}
\bibinfo{author}{G.~Caimi}, \bibinfo{author}{F.~Fischer}, \bibinfo{author}{T.~Schlechte}, \bibinfo{title}{{Railway Track Allocation}}, volume \bibinfo{volume}{268}, \bibinfo{year}{2018}. \DOIprefix\doi{10.1007/978-3-319-72153-8{\_}7}.
\bibitem[{Shao et~al.(2022)Shao, Xu, Sun, Kong, and Lu}]{Shao2022Equity-orientedSystem}
\bibinfo{author}{J.~Shao}, \bibinfo{author}{Y.~Xu}, \bibinfo{author}{L.~Sun}, \bibinfo{author}{D.~Kong}, \bibinfo{author}{H.~Lu},
\newblock \bibinfo{title}{{Equity-oriented integrated optimization of train timetable and stop plans for suburban railways system}},
\newblock \bibinfo{journal}{Computers {\&} Industrial Engineering} \bibinfo{volume}{173} (\bibinfo{year}{2022}) \bibinfo{pages}{108721}. \URLprefix \url{https://www.sciencedirect.com/science/article/pii/S0360835222007094}. \DOIprefix\doi{10.1016/J.CIE.2022.108721}.
\bibitem[{Gestrelius et~al.(2020)Gestrelius, Peterson, and Aronsson}]{Gestrelius2020}
\bibinfo{author}{S.~Gestrelius}, \bibinfo{author}{A.~Peterson}, \bibinfo{author}{M.~Aronsson},
\newblock \bibinfo{title}{{Timetable quality from the perspective of a railway infrastructure manager in a deregulated market: An interview study with Swedish practitioners}},
\newblock \bibinfo{journal}{Journal of Rail Transport Planning and Management} \bibinfo{volume}{15} (\bibinfo{year}{2020}). \DOIprefix\doi{10.1016/j.jrtpm.2020.100202}.
\bibitem[{Cao and Feng(2020)}]{Cao2020}
\bibinfo{author}{C.~Cao}, \bibinfo{author}{Z.~Feng},
\newblock \bibinfo{title}{{Optimal capacity allocation under random passenger demands in the high-speed rail network}},
\newblock \bibinfo{journal}{Engineering Applications of Artificial Intelligence} \bibinfo{volume}{88} (\bibinfo{year}{2020}). \DOIprefix\doi{10.1016/j.engappai.2019.103363}.
\bibitem[{Bruzzone et~al.(2023)Bruzzone, Cavallaro, and Nocera}]{Bruzzone2023}
\bibinfo{author}{F.~Bruzzone}, \bibinfo{author}{F.~Cavallaro}, \bibinfo{author}{S.~Nocera},
\newblock \bibinfo{title}{{The definition of equity in transport}},
\newblock in: \bibinfo{booktitle}{Transportation Research Procedia}, volume~\bibinfo{volume}{69}, \bibinfo{year}{2023}, pp. \bibinfo{pages}{440--447}. \DOIprefix\doi{10.1016/j.trpro.2023.02.193}.
\bibitem[{Li et~al.(2019)Li, Zhang, Dong, Yin, and Cao}]{Zhang2019}
\bibinfo{author}{D.~Li}, \bibinfo{author}{T.~Zhang}, \bibinfo{author}{X.~Dong}, \bibinfo{author}{Y.~Yin}, \bibinfo{author}{J.~Cao},
\newblock \bibinfo{title}{{Trade-off between efficiency and fairness in timetabling on a single urban rail transit line under time-dependent demand condition}},
\newblock \bibinfo{journal}{Transportmetrica B} \bibinfo{volume}{7} (\bibinfo{year}{2019}) \bibinfo{pages}{1203--1231}. \DOIprefix\doi{10.1080/21680566.2019.1589598}.
\bibitem[{van Wee and Mouter(2021)}]{Wee2021}
\bibinfo{author}{B.~van Wee}, \bibinfo{author}{N.~Mouter},
\newblock \bibinfo{title}{{Chapter Five - Evaluating transport equity}},
\newblock in: \bibinfo{editor}{N.~Mouter} (Ed.), \bibinfo{booktitle}{Advances in Transport Policy and Planning}, volume~\bibinfo{volume}{7}, \bibinfo{publisher}{Academic Press}, \bibinfo{year}{2021}, pp. \bibinfo{pages}{103--126}. \URLprefix \url{https://www.sciencedirect.com/science/article/pii/S2543000920300354}. \DOIprefix\doi{https://doi.org/10.1016/bs.atpp.2020.08.002}.
\bibitem[{Zhang et~al.(2024)Zhang, Li, Yuan, Zhang, and Yang}]{Zhang2024}
\bibinfo{author}{Y.~Zhang}, \bibinfo{author}{S.~Li}, \bibinfo{author}{Y.~Yuan}, \bibinfo{author}{J.~Zhang}, \bibinfo{author}{L.~Yang},
\newblock \bibinfo{title}{{Approximate dynamic programming approach to efficient metro train timetabling and passenger flow control strategy with stop-skipping}},
\newblock \bibinfo{journal}{Engineering Applications of Artificial Intelligence} \bibinfo{volume}{127} (\bibinfo{year}{2024}). \DOIprefix\doi{10.1016/j.engappai.2023.107393}.
\bibitem[{Karsu and Morton(2015)}]{Karsu2015}
\bibinfo{author}{{\"{O}}.~Karsu}, \bibinfo{author}{A.~Morton},
\newblock \bibinfo{title}{{Inequity averse optimization in operational research}},
\newblock \bibinfo{journal}{European Journal of Operational Research} \bibinfo{volume}{245} (\bibinfo{year}{2015}) \bibinfo{pages}{343--359}. \DOIprefix\doi{10.1016/j.ejor.2015.02.035}.
\bibitem[{Luan et~al.(2023)Luan, Sun, Corman, and Meng}]{Luan2023}
\bibinfo{author}{X.~Luan}, \bibinfo{author}{X.~Sun}, \bibinfo{author}{F.~Corman}, \bibinfo{author}{L.~Meng},
\newblock \bibinfo{title}{{Inequity averse optimization of railway traffic management considering passenger route choice and Gini Coefficient}},
\newblock \bibinfo{journal}{Journal of Rail Transport Planning and Management} \bibinfo{volume}{26} (\bibinfo{year}{2023}). \DOIprefix\doi{10.1016/j.jrtpm.2023.100395}.
\bibitem[{Reynolds et~al.(2023)Reynolds, Ehrgott, and Wang}]{Reynolds2023}
\bibinfo{author}{E.~Reynolds}, \bibinfo{author}{M.~Ehrgott}, \bibinfo{author}{J.~Wang},
\newblock \bibinfo{title}{{An evaluation of the fairness of railway timetable rescheduling in the presence of competition between train operators}},
\newblock \bibinfo{journal}{Journal of Rail Transport Planning and Management} \bibinfo{volume}{26} (\bibinfo{year}{2023}). \DOIprefix\doi{10.1016/j.jrtpm.2023.100389}.
\bibitem[{Aron and Abraham(2022)}]{Aron2022}
\bibinfo{author}{R.~Aron}, \bibinfo{author}{A.~Abraham},
\newblock \bibinfo{title}{{Resource scheduling methods for cloud computing environment: The role of meta-heuristics and artificial intelligence}},
\newblock \bibinfo{journal}{Engineering Applications of Artificial Intelligence} \bibinfo{volume}{116} (\bibinfo{year}{2022}). \DOIprefix\doi{10.1016/j.engappai.2022.105345}.
\bibitem[{Jain et~al.(1984)Jain, Chiu, and Hawe}]{Jain1998}
\bibinfo{author}{R.~K. Jain}, \bibinfo{author}{D.-M.~W. Chiu}, \bibinfo{author}{W.~R. Hawe},
\newblock \bibinfo{title}{{A quantitative measure of fairness and discrimination}},
\newblock \bibinfo{journal}{Eastern Research Laboratory, Digital Equipment Corporation, Hudson, MA} \bibinfo{volume}{21} (\bibinfo{year}{1984}).
\bibitem[{Gini(1912)}]{Gini1912}
\bibinfo{author}{C.~Gini}, \bibinfo{title}{{Variabilit{\`{a}} e mutabilit{\`{a}}: contributo allo studio delle distribuzioni e delle relazioni statistiche.[Fasc. I.]}}, \bibinfo{publisher}{Tipogr. di P. Cuppini}, \bibinfo{year}{1912}.
\bibitem[{Atkinson(1970)}]{Atkinson1970}
\bibinfo{author}{A.~B. Atkinson},
\newblock \bibinfo{title}{{On the measurement of inequality}},
\newblock \bibinfo{journal}{Journal of economic theory} \bibinfo{volume}{2} (\bibinfo{year}{1970}) \bibinfo{pages}{244--263}.
\bibitem[{Abd-Elnaby(2021)}]{Elnaby2021}
\bibinfo{author}{M.~Abd-Elnaby},
\newblock \bibinfo{title}{{Capacity and fairness maximization-based resource allocation for downlink NOMA networks}},
\newblock \bibinfo{journal}{Computers, Materials and Continua} \bibinfo{volume}{69} (\bibinfo{year}{2021}) \bibinfo{pages}{521--537}. \DOIprefix\doi{10.32604/cmc.2021.018351}.
\bibitem[{Avcil et~al.(2024)Avcil, Soyturk, and Kantarci}]{Avcil2024}
\bibinfo{author}{M.~Avcil}, \bibinfo{author}{M.~Soyturk}, \bibinfo{author}{B.~Kantarci},
\newblock \bibinfo{title}{{Fair and efficient resource allocation via vehicle-edge cooperation in 5G-V2X networks}},
\newblock \bibinfo{journal}{Vehicular Communications} \bibinfo{volume}{48} (\bibinfo{year}{2024}). \DOIprefix\doi{10.1016/j.vehcom.2024.100773}.
\bibitem[{Romero et~al.(2016)Romero, Nozick, and Xu}]{Romero2016}
\bibinfo{author}{N.~Romero}, \bibinfo{author}{L.~Nozick}, \bibinfo{author}{N.~Xu},
\newblock \bibinfo{title}{{Hazmat facility location and routing analysis with explicit consideration of equity using the Gini coefficient}},
\newblock \bibinfo{journal}{Transportation Research Part E: Logistics and Transportation Review} \bibinfo{volume}{89} (\bibinfo{year}{2016}) \bibinfo{pages}{165--181}. \DOIprefix\doi{10.1016/j.tre.2016.02.008}.
\bibitem[{Raza et~al.(2023)Raza, Akuh, Safdar, and Zhong}]{Raza2023}
\bibinfo{author}{A.~Raza}, \bibinfo{author}{R.~Akuh}, \bibinfo{author}{M.~Safdar}, \bibinfo{author}{M.~Zhong},
\newblock \bibinfo{title}{{Public transport equity with the concept of time-dependent accessibility using Geostatistics methods, Lorenz curves, and Gini coefficients}},
\newblock \bibinfo{journal}{Case Studies on Transport Policy} \bibinfo{volume}{11} (\bibinfo{year}{2023}). \DOIprefix\doi{10.1016/j.cstp.2023.100956}.
\bibitem[{Villa-Zapata et~al.(2024)Villa-Zapata, Rodriguez-Roman, Fl{\'{o}}rez-Coronel, Gonz{\'{a}}lez-L{\'{o}}pez, and Figueroa-Medina}]{Lina2024}
\bibinfo{author}{L.~M. Villa-Zapata}, \bibinfo{author}{D.~Rodriguez-Roman}, \bibinfo{author}{J.~E. Fl{\'{o}}rez-Coronel}, \bibinfo{author}{J.~M. Gonz{\'{a}}lez-L{\'{o}}pez}, \bibinfo{author}{A.~M. Figueroa-Medina},
\newblock \bibinfo{title}{{Incorporating equity in the vehicle rebalancing operations of dockless micromobility services}},
\newblock \bibinfo{journal}{Latin American Transport Studies} \bibinfo{volume}{2} (\bibinfo{year}{2024}) \bibinfo{pages}{100009}.
\bibitem[{Chen et~al.(2025)Chen, Ki, Li, and Wang}]{Chen2025}
\bibinfo{author}{Z.~Chen}, \bibinfo{author}{D.~Ki}, \bibinfo{author}{Z.~Li}, \bibinfo{author}{K.~Wang},
\newblock \bibinfo{title}{{Assessing equity in infrastructure investment distribution among U.S. cities}},
\newblock \bibinfo{journal}{Cities} \bibinfo{volume}{162} (\bibinfo{year}{2025}) \bibinfo{pages}{105898}. \DOIprefix\doi{10.1016/J.CITIES.2025.105898}.
\bibitem[{Holland(1992)}]{Holland1992}
\bibinfo{author}{J.~H. Holland}, \bibinfo{title}{{Adaptation in natural and artificial systems: an introductory analysis with applications to biology, control, and artificial intelligence}}, \bibinfo{publisher}{MIT press}, \bibinfo{year}{1992}.
\bibitem[{Van~Thieu and Mirjalili(2023)}]{Vanthieu2023}
\bibinfo{author}{N.~Van~Thieu}, \bibinfo{author}{S.~Mirjalili},
\newblock \bibinfo{title}{{MEALPY: An open-source library for latest meta-heuristic algorithms in Python}},
\newblock \bibinfo{journal}{Journal of Systems Architecture} \bibinfo{volume}{139} (\bibinfo{year}{2023}). \DOIprefix\doi{10.1016/j.sysarc.2023.102871}.

\end{thebibliography}

\end{document}